\documentclass[manuscript,nonacm]{acmart} 

\usepackage{multirow}
\usepackage{amsmath}
\usepackage{graphicx}
\usepackage{array}
\usepackage{adjustbox}
\usepackage{tabularx}
\usepackage{float}
\usepackage{longtable}
\usepackage{geometry}
\usepackage{amssymb}
\usepackage{placeins}  
\AtBeginDocument{%
  \providecommand\BibTeX{{\normalfont B\kern-0.5em{\scshape i\kern-0.25em b}\kern-0.8em\TeX}}}

\renewcommand\footnotetextcopyrightpermission[1]{} 

\begin{document}

\title{Quantum Natural Language Processing: A Comprehensive Review of Models, Methods, and Applications}

\author{Farha Nausheen}
\email{farha.nausheen@live.vu.edu.au}
\orcid{0000-0002-0112-9354}
\affiliation{%
  \institution{Victoria University}
  \city{Melbourne}
  \state{Victoria}
  \country{Australia}
}

\author{Khandakar Ahmed}
\email{khandakar.ahmed@vu.edu.au}
\orcid{}
\affiliation{%
  \institution{Victoria University}
  \city{Melbourne}
  \state{Victoria}
  \country{Australia}
}
\author{M Imad Khan}
\email{muhammed.khan@vu.edu.au}
\affiliation{%
  \institution{Victoria University}
  \city{Melbourne}
  \state{Victoria}
  \country{Australia}
}
\author{Farina Riaz}
\email{farina.riaz@csiro.au}
\affiliation{%
  \institution{CSIRO}
  \city{Sydney}
  \state{NSW}
  \country{Australia}
}

\begin{abstract}
In recent developments, deep learning methodologies applied to Natural Language Processing (NLP) have revealed a paradox: They improve performance but demand considerable data and resources for their training. Alternatively, quantum computing exploits the principles of quantum mechanics to overcome the computational limitations of current methodologies, thereby establishing an emerging field known as quantum natural language processing (QNLP). This domain holds the potential to attain a quantum advantage in the processing of linguistic structures, surpassing classical models in both efficiency and accuracy. In this paper, it is proposed to categorise QNLP models based on quantum computing principles, architecture, and computational approaches.This paper attempts to provide a survey on how quantum meets language by mapping state-of-the-art in this area, embracing quantum encoding techniques for classical data, QNLP models for prevalent NLP tasks, and quantum optimisation techniques for hyper parameter tuning. The landscape of quantum computing approaches applied to various NLP tasks is summarised by showcasing the specific QNLP methods used, and the popularity of these methods is indicated by their count. From the findings, it is observed that QNLP approaches are still limited to small data sets, with only a few models explored extensively, and there is increasing interest in the application of quantum computing to natural language processing tasks.
\end{abstract}

\maketitle
\section{INTRODUCTION}
Natural Language Processing (NLP) enables the interpretation and analysis of human language by computational systems through the integration of principles derived from Artificial Intelligence (AI), Linguistics, and Computer Science. Recent advances in deep learning-driven natural language processing (NLP) have resulted in notable performance gains. However, these improvements have been accompanied by an escalation in computational complexity, a heightened demand for extensive datasets, and substantial resource utilisation. The reliance on expansive neural networks for numerous NLP tasks comes with notable limitations, such as being energy-inefficient, struggling to generalise with limited data, and facing computational constraints. Quantum computing (QC) has emerged as a potential solution to these challenges, serving as an accelerator for AI and NLP. 
Using quantum mechanical principles, QC offers a new dimension of computational power by drastically accelerating the processing of complex and resource-intensive problems\cite{10577116}. In particular, Quantum Natural Language Processing (QNLP)\cite{coecke2020foundationsneartermquantumnatural} represents an emerging field that amalgamates quantum principles with methodologies from natural language processing. Its objectives include improving model efficiency, optimising computational complexity, and establishing novel paradigms for linguistic representations. Unlike traditional NLP models that require extensive parametrisation, QNLP approaches harness quantum states to encode linguistic structures, enabling compact and entangled representations that could lead to more efficient learning mechanisms.

QNLP models capitalise on the unique expressive power of quantum computing, facilitating reduced memory overhead, enhanced generalisation from small datasets, and improved optimisation of NLP tasks. Hybrid quantum-classical algorithms have been formulated for tasks such as
sentence generation, employing techniques like simulated annealing to address combinatorial optimisation challenges, thereby
demonstrating the prospective utility of near-term quantum devices in the domain of NLP applications. 
\cite{karamlou2022quantumnaturallanguagegeneration}. Additionally, Variational Quantum Algorithms (VQAs) serve as learning components to dynamically capture and analyse the syntactic and semantic aspects of text. This facilitates improving NLP efficiency, offering alternative approaches to feature extraction, text encoding, and language understanding. 
This survey presents a systematic review of the existing literature on Quantum NLP, focusing on:
\begin{itemize}
    \item	Quantum encoding techniques for classical data, exploring how NLP inputs can be efficiently mapped into quantum states.
    \item QNLP models and their applications, classification, and analysis of various frameworks designed for NLP tasks.
    \item Quantum optimisation techniques, discussing their role in hyperparameter tuning and computational efficiency improvements.
    \item The landscape of QNLP research, examining the adoption trends of different quantum-based NLP methodologies.
\end{itemize}
By consolidating existing research and delineating current limitations, this review seeks to illuminate prospective advances in quantum natural language processing (QNLP), with particular emphasis on overcoming challenges related to scalability, efficiency, and practical implementation. The findings reveal that QNLP is still an emerging field, with models remaining restricted by limited datasets and nascent quantum hardware technology. However, increasing interest in quantum-enhanced NLP techniques suggests a promising path to achieve quantum advantage in language processing.
The structural organisation of this article is delineated as follows. Section 2 provides an exposition of the related work, emphasising significant contributions. Section 3 outlines the influences that have inspired the research that culminated in this survey. Section 4 presents the survey methodology detailing the planning, execution, and reporting phase of the review process. The information in Section 5 covers the reporting and analysis of the results of the survey questions. Section 6 elaborates on the findings, discussion, and limitations inherent in this study. Section 7 concludes the manuscript by delineating the concluding remarks, summarising the contributions, and illuminating prospective directions for future research.

\section{RELATED WORK}
The fundamental principles of quantum computing, specifically superposition and entanglement, are crucial for the application of Quantum Natural Language Processing (QNLP) in the manipulation of linguistic data. While the potential of quantum computing to fundamentally reshape the field of natural language processing is substantial, its current advancement remains reliant on classical computational resources and is impeded by the nascent stage of quantum hardware development. Lambeq\cite{kartsaklis2021lambeqefficienthighlevelpython}, an open source modular Python toolkit [3] offers modules and classes needed to transform a sentence into a quantum circuit, is available to experiment with quantum natural language processing.
The present generation of quantum computing systems is referred to as Noisy Intermediate-Scale Quantum (NISQ) devices\cite{Preskill_2018}.These devices comprise around hundreds of qubits and are characterised by significant noise during computations. NISQ devices are particularly vulnerable to decoherence, which is the loss of information in a qubit due to unwanted interactions with the environment. This decoherence restricts the maximum depth of circuits that can be executed on these devices. Despite these limitations, Google has provided evidence of quantum supremacy using a 53-qubit quantum computer\cite{arute2019quantum}.
By applying the principles of quantum computing, the DisCoCat framework allows for the conception of natural language processing tasks as quantum computations, leading to a quadratic speed-up in calculating sentence similarity. Furthermore, the focus is on the implementation of a grammar-sensitive question-answering system\cite{Meichanetzidis_2023} on quantum hardware, achieved through the mapping of grammatical sentences onto parameterised quantum circuits (PQC). 
Grammar aware sentence classification \cite{meichanetzidis2023grammar} employs DisCoCat framework to instantiate sentences and word meanings as parameterised quantum circuits and quantum state respectively. The grammatical structure of a sentence is meticulously represented as a configuration of entangling operations, which integrate individual word circuits into an integrated sentence circuit. Quantum Natural Language Processing (QNLP) for sentence classification\cite{10.1007/978-3-031-35644-5_7}, focuses on small data sets and simple sentence structure, and shows promising results with COMPASS optimisation, especially with varying ansatz depths.
\\A sentiment classification methodology based on machine learning has been developed for the Arabic language, specifically tailored for implementation on quantum computers. This methodology enables a comparative analysis between quantum computing and machine learning techniques for the classification of Arabic-language documents, utilising two distinct data sets\cite{omar2023quantum}. An alternative approach for sentiment analysis involves employing complex-valued embeddings inspired by quantum measurements alongside the quantum long-short-term memory network\cite{chu2024effective}. This method is designed to effectively identify interactions between sentences and accurately capture contextual semantic details. Quantum natural language processing-based sentiment analysis using the lambeq toolkit \cite{ganguly2022quantum} utilises the lambeq toolkit on intermediate book genre sentiments dataset.

A simple translation from English to Urdu grammar uses DisCoCirc framework \cite{Waseem_2022} to eliminate grammatical differences between languages. Quantum machine translation uses the DisCoCat framework to represent sentences in two syntactically different languages\cite{abbaszade2023quantummachinetranslationsyntactically}. QNLP in practice\cite{lorenz2023qnlp} discusses the implementation of compositional models of meaning on quantum hardware, demonstrating how sentence representations can be mapped to quantum circuits. Variational quantum algorithms (VQA) \cite{cerezo2021variational} employ a classical optimiser to optimise parameters and provide noise resistance, thus improving computational efficiency. 
Quantum neural networks constitute a class of machine learning models precisely designed for implementation on quantum computing architectures. Quantum principles, including superposition, entanglement, and interference, are employed to execute computations. The capabilities of quantum neural networks \cite{abbas2021power} exhibit a remarkable Fisher information spectrum, which facilitates accelerated training compared to both classical and quantum models. The Quantum Self-Attention Neural Network \cite{Zeng_2016} constitutes a Gaussian-projected quantum self-attention architecture, devised to address the constraints of substantial syntactic pre-processing in QNLP, and is pertinent for near-term quantum devices. Another Quantum Self-Attention Mechanism (QSAM) \cite{shi2023naturalnisqmodelquantum} delineates fundamental operations, including the computation of attention scores and features, by employing the data encoding and ansatz architecture in an appropriate manner. QMSAN \cite{chen2024quantummixedstateselfattentionnetwork} employs a quantum attention mechanism based on mixed states. This approach augments the expressive capacity of the quantum system by utilising mixed-state operations, thereby facilitating the model's ability to more efficiently and accurately discern the similarities between queries and keys.
POVM-QSANN \cite{Wei2023PovmBasedQS} constitutes a positive operator-valued measure (POVM) approach that employs operators to extract a wide range of information from qubits. This method significantly improves the capacity for information extraction and optimally uses feature space.
Paraphrase Identification using DisCoCat Model\cite{khatriexperimental}, utilises born rule, simply prepares states for the 2 sentences to be compared, and evaluates the diagram through tensor contraction. The square norm of the resulting complex number is then calculated to ensure that we get a real value between 0 and 1.
\section{RESEARCH MOTIVATION}
In this section, the necessary background information is presented to understand the motivation and significance of this survey.
\subsection{Background}Lambeq is a sophisticated quantum software toolkit explicitly created for quantum natural language processing (QNLP), utilising categorical compositional models to transform linguistic structures into quantum circuits. Conversely, quantum neural networks (QNNs) for natural language processing employ parameterised quantum circuits to acquire representations and execute downstream language tasks, such as classification, frequently drawing inspiration from classical deep learning paradigms.
\subsubsection{Lambeq}
\mbox{} \\
Lambeq\cite{kartsaklis2021lambeqefficienthighlevelpython}, developed in 2021 by a research team at Cambridge Quantum Computing, constitutes an open-source, modular, and extensible high-level Python library. It provides a suite of tools for the implementation of experimental Quantum Natural Language Processing (QNLP), facilitating the transformation of sentences into string diagrams, tensor networks, and quantum circuits suitable for execution on a quantum computer. Applications of lambeq encompass text classification \cite{10.1007/978-3-031-28540-0_17}, \cite{lorenz2023qnlp}, sentiment classification \cite{9951286}, \cite{ganguly2022quantum}, among several others.
\begin{figure}[ht]
    \centering
\includegraphics[width=0.8\textwidth]{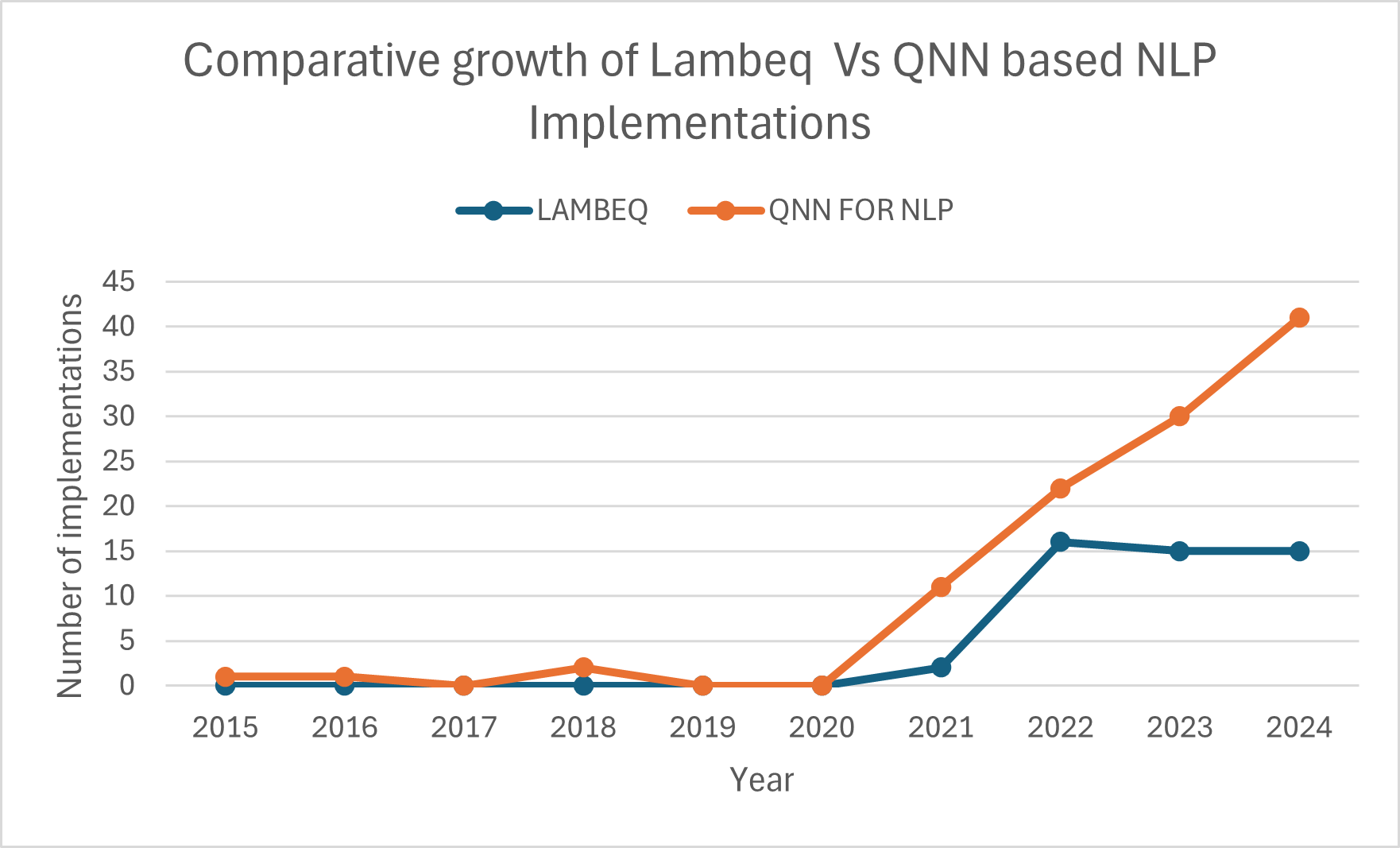}  
    \caption{QNLP Adoption by lambeq and QNN models}
\label{fig:LambeqQNNTrend}
\end{figure}

\subsubsection{QML for NLP}
\mbox{} \\
Quantum Machine Learning (QML) represents a progressive domain that integrates quantum computing with traditional machine learning (ML) methodologies, aiming to enhance computational complexity by exceeding the capabilities of classical computing systems. The contemporary generation of quantum apparatuses, referred to as Noisy Intermediate-Scale Quantum (NISQ) devices, encompasses up to several hundred qubits. Although these devices experience quantum noise due to imperfections and fluctuations in the quantum gates, they have demonstrated a quantum advantage in certain tasks by outperforming classical computers\cite{correia2022groversalgorithmquestionanswering}. QML has the potential to revolutionise AI systems by exploiting quantum neural networks (QNN) \cite{abbas2021power}. These networks have demonstrated promising results in domains such as classification, sentiment analysis, language translation
In spite of these considerable advancements, the field of Quantum Machine Learning (QML) within the Noisy Intermediate-Scale Quantum (NISQ) era is still in its formative stage. As quantum hardware continues to develop, there is strong motivation to formulate more efficient QML algorithms that can fully leverage quantum capabilities, thereby establishing a foundation for future artificial intelligence systems that exploit the unique benefits of quantum computing.
\subsubsection{Lambeq Vs QML}
\mbox{} \\
The comparative analysis of the adoption trends of the Lambeq toolkit, as opposed to Quantum Neural Network (QNN)-based natural language processing (NLP) implementations from 2015 to 2024, as illustrated in Figure ~\ref{fig:LambeqQNNTrend} reveals a notable divergence.
The data in figure come from a manual bibliometric analysis using Google Scholar, which estimates yearly implementations (2015-2024) using specific keywords for Lambeq and QNN-based models. The analysis focused on works detailing practical implementations, only considering relevant peer-reviewed papers or pre prints. Relevance was manually verified to show trends in Quantum NLP adoption.
It can be discerned from these implementations that Lambeq exhibited initial growth from the year 2021 to the year 2022, reflecting approximately 15 implementations. However, beginning in 2022, the growth rate experienced a decline, marked by merely a slight increase in 2024. This slowdown can be ascribed to researchers incorporating machine learning techniques alongside DisCoCat-based Lambeq implementations, leading to a shift towards hybrid models instead of the singular utilisation of Lambeq.
On the other hand, QNN for NLP implementations demonstrates continuous growth from the year 2022 to 2024. The deployment of QNN methodologies has progressed at a consistent rate, exceeding that of Lambeq in 2022 and exhibiting a pronounced increase through 2024. By 2024, the number of QNN-based applications had more than doubled compared to Lambeq, indicative of an increasing preference for quantum NLP strategies that are underpinned by deep learning paradigms.
Most of these Lambeq implementations, even while pioneering, lack scalability because they rely on syntactic analysis, a pre processing task that requires significant effort, particularly with large datasets. Additionally, these syntax-based
methods are not very flexible for handling the wide range of complex expressions found in human language for sentences with varied syntactical structures. Within the sphere of Natural Language Processing (NLP), Quantum Neural Networks (QNNs) are undergoing rapid evolution because of their flexibility, ability to scale, and effective integration with classic deep learning techniques. Hybridisation of QNLP models yields the benefits of quantum computing alongside classical approaches, enhancing both quantum and machine learning capabilities. 
\subsection{Comparison with existing surveys}
QNLP represents a nascent field, where existing research surveys provide valuable information predominantly focused on fundamental theoretical frameworks and limited-scale implementations, in contrast to extensive surveys.  The survey \cite{guarasci2022quantum} systematically categorises existing methodologies based on their typology, including theoretical frameworks or tangible applications on classical or quantum platforms. It further delineates these approaches according to task, distinguishing between generalised syntax-semantic representation and specific NLP tasks such as sentiment analysis and question answering. Additionally, it assesses the evaluative resources utilised, whether they are standard benchmarks or custom-designed datasets.\cite{10577116} provides an exhaustive survey on quantum natural language processing by identifying models such as the quantum bag-of-words model, the TPR model, the DisCoCat model, as well as the word2ket and word2ketXS models. \cite{Widdows_2024} surveys quantum natural language processing, scrutinising natural language processing techniques such as word embeddings, sequential models, syntactic parsing, and attention mechanisms within transformer architectures, while proposing an innovative quantum-based text encoding paradigm. Furthermore, it delves into the implications of quantum theory in elucidating concepts of uncertainty and intelligence. \cite{wu2021natural} investigates notable techniques at the confluence of NLP and quantum physics over the preceding decade, systematically classifying them based on quantum theory, linguistic objectives, and subsequent applications.
These surveys provide significant insight into the dynamic discipline of QNLP that addresses key methodologies, model classifications, and quantum-enhanced NLP techniques. However, existing research in quantum computing's applicability to diverse NLP tasks remains substantially under explored, underscoring the critical need for continued scholarly inquiry within this domain. Table 1 provides a comparative analysis of multiple surveys conducted between 2021 and 2024, in alignment with the focus of the present survey.\\
To the best of our understanding, this survey offers innovative contributions in the following domains, setting it apart from preceding QNLP surveys.
\begin{itemize}
    \item Although previous surveys frequently restricted themselves to summarising existing literature, our study progresses further by categorising and analysing QNLP models and their applications. We present a taxonomy of frameworks specifically designed for a variety of NLP tasks.
    \item In this study, we seek to align quantum natural language processing (QNLP) models with natural language processing (NLP) tasks such as sentiment analysis, question answering, text summarisation, and language translation. This alignment aims to determine which quantum models exhibit optimal performance for each specific NLP challenge.
    \item We analyse quantum encoding methods for classical datasets, focusing on transforming Natural Language Processing inputs into quantum states, to connect theory with practical applications.
    \item We explore quantum optimisation techniques, with a specific focus on their use in hyperparameter tuning.
In conclusion, this survey provides a comprehensive landscape analysis of adoption trends in Quantum Natural Language Processing (QNLP) research. By doing so, it not only disseminates information, but also situates the progression of quantum NLP within a broader context, thereby laying a robust groundwork for future advancements in the field.
\end{itemize}

\begin{table}[ht]  
\centering
\normalsize 
\renewcommand{\arraystretch}{1.5} 
\setlength{\tabcolsep}{4pt} 

\caption{Comparative analysis of various surveys conducted between 2021 and 2024 in relation to the present survey}

\resizebox{\textwidth}{!}{%
\begin{tabular}{|>{\centering\arraybackslash}p{1.8cm}|
                >{\centering\arraybackslash}p{1.2cm}|
                >{\centering\arraybackslash}p{2.5cm}|
                >{\centering\arraybackslash}p{2.5cm}|
                >{\centering\arraybackslash}p{2.5cm}|
                >{\centering\arraybackslash}p{2.5cm}|
                >{\centering\arraybackslash}p{3.2cm}|
                >{\centering\arraybackslash}p{3.2cm}|}
\hline
\textbf{Ref.} & \textbf{Year} & \textbf{QNLP Applications} & \textbf{Quantum Encoding} & \textbf{Quantum Optimisation} & \textbf{Quantum Algorithms} & \textbf{Categories of QNLP Models} & \textbf{Practical Implementations} \\
\hline
\cite{wu-etal-2021-natural} & 2021 & \checkmark & \checkmark & \checkmark & \checkmark & -- & \checkmark \\
\cite{guarasci2022quantum} & 2022 & \checkmark & $\Delta$ & $\Delta$ & \checkmark & -- & $\Delta$ \\
\cite{10577116} & 2024 & \checkmark & \checkmark & \checkmark & \checkmark & -- & $\Delta$ \\
\cite{Widdows_2024} & 2024 & \checkmark & \checkmark & \checkmark & \checkmark & -- & \checkmark \\
\textit{This Survey} & 2025 & \checkmark & \checkmark & \checkmark & \checkmark & \checkmark & \checkmark \\
\hline
\end{tabular}
}
\begin{flushleft}
\footnotesize The \checkmark{} symbol indicates extensive coverage, while $\Delta$ denotes areas that were addressed but could be expanded. “--” indicates areas not explicitly covered.
\label{table:SurveysComparison}
\end{flushleft}
\end{table}
\section{REVIEWING METHODOLOGY}
\begin{figure}[ht]
  \centering
  \includegraphics[width=0.8\textwidth]{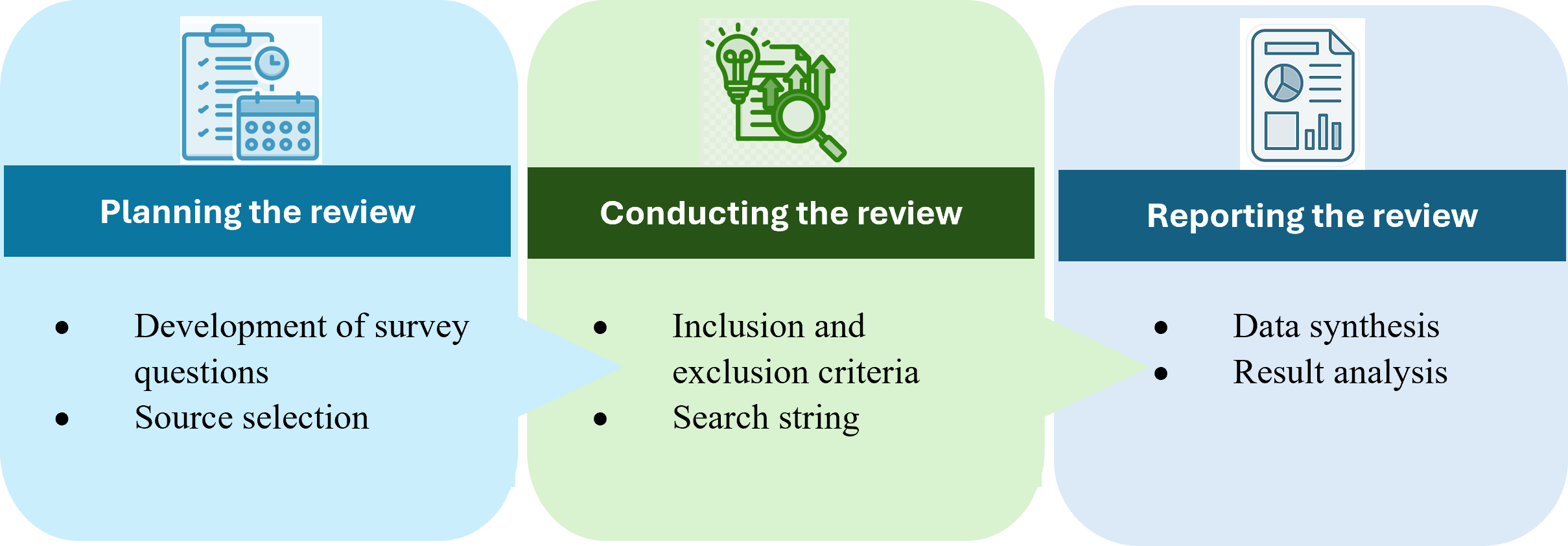}
  \caption{Review methodology of the SLR stages}
  \label{fig:review methodology}
  \end{figure}

In order to systematically identify, evaluate, and interpret the literature in response to the proposed survey questions, the methodology outlined by the Kitchenam\cite{kitchenham2004procedures} guideline is utilised for the planning, execution, and review of the study. Each phase consists of one or more specific steps. This framework has been adopted and developed a methodological workflow comprising nine processing steps within these three phases, providing greater flexibility for future modifications of the SLR. The key phases of the SLR are illustrated in Figure~\ref{fig:review methodology}.
The structure of this section is organised as \cite{kitchenham2007guidelines}. First, databases will be chosen, followed by the formulation of survey questions and the establishment of inclusion and exclusion criteria. The search string will then be defined and the review process will be conducted.
\subsection{SLR Planning Phase}
This review seeks to address solutions and advancements in quantum natural language processing, quantum encoding techniques, and quantum optimisation mechanisms. The objective of this paper is to provide an introduction to the domain of quantum natural language processing, to examine recent advancements in the field, and to identify existing challenges.

With the recent advent of the quantum computing discipline, this review will concentrate on advancements and applications spanning the years 2012 to 2024.The review will discuss various techniques and explore the related challenges. Articles for this review were retrieved from the ------. The criteria impacting the consideration of these databases were relevance to computer science research, broad article collection, and accessibility through Victoria University's licences. The following phases are planned for conducting the review:

1. Formulation of the survey questions 

2. Source selection

3. Inclusion and exclusion criteria

4. Search strings.

\subsubsection{Development of Survey Questions}
The objective of the survey is to examine the prevailing condition of natural language processing in the quantum computing environment. The survey questions are structured to encompass all the aspects involving quantum computing for natural language processing comprehensively.

SQ 1:How can quantum algorithms improve the efficiency of natural language processing tasks?
\\
\textit{Justification: What are the theoretical foundations of QNLP and how do they differ from classical NLP approaches?}
\\
SQ 2: How can classical text data be effectively encoded into quantum states for use in quantum machine learning models?\\
\textit{Justification: How can text or information be represented as quantum states and what are the different types of embedding used in quantum computing?}
\\
SQ 3: What are the general categories of Quantum Natural Language Processing(QNLP) models? \\
\textit{Justification: Implementing quantum natural language processing involves several innovative approaches that exploit quantum computing principles }
\\
SQ 4: What are the different NLP tasks handled using quantum computing?\\
\textit{Justification: What are the various methodologies through which NLP tasks can be addressed using quantum techniques?}\\
SQ 5: In what ways might quantum optimisation algorithms be employed to enhance hyper-parameter tuning or other optimisation tasks within natural language processing models?\\
\textit{Justification} How Can Quantum Optimisation Algorithms Enhance Hyper-Parameter Tuning in NLP?
\subsubsection{Source selection}
To define an efficient data extraction strategy, a selection of research sources (e.g. Science Direct, ACM Digital Library, Google Scholar, IBM Quantum papers and other sources) was evaluated based on their relevance to the specified area of research.  The selection of journals and conference papers was deliberately limited to those that make significant contributions to the field of quantum natural language processing.  Various strategies were used to meticulously search through different libraries to identify pertinent content. In addition, references in the bibliographies of the selected articles with respect to the selected area of interest were scrutinised to augment the repository of relevant sources.
\subsection {SLR Conducting Phase}
This SLR aims to illustrate research that uses quantum computing for natural language processing(NLP) tasks. The conduct phase will involve tasks such as defining inclusion, exclusion criteria, and search strings. 

\subsubsection{Inclusion and Exclusion criteria}
To obtain improved performance for the evaluation of targeted scholarly works, specific criteria were established. Inclusion criteria (IC) must be fully met for the work to be considered as part of search, whereas the Exclusion criteria (EC) are used to exclude works from the search. 

IC 1: Articles that use quantum computing on natural language processing. (AND)

IC 2: Articles proposing a QNLP solution. (AND)

IC 3: Articles are published in journals, books, pre-printed or conferences. (AND)

IC 4: Articles are published in English.

EC 1: Articles that do not satisfy the IC conditions. (OR) 

EC 2: Articles are not in a Journal, Book, Pre-Print, or Conference. 

\subsubsection{Search string/query string }
The initial step involves collecting all articles from 2012 to 2024. This time frame was chosen because it aligns with the emergence and expansion of quantum devices, as well as the development and possible use of noisy intermediate-scale quantum (NISQ) devices \cite {Preskill_2018} To facilitate this process, we have identified some keywords: quantum natural language processing (QNLP), Quantum Neural Networks, DisCoCat, Sentiment Analysis, Text Classification, Text Summarisation, Language Translation, Question Answering.   
\subsection{SLR Reporting Phase}
To conduct the report phase, two tasks are performed, namely {\itshape data synthesis} and {\itshape result analysis}. The results are summarised in the literature review execution step, is an essential component of this methodology \cite{kitchenham2004procedures}. In the final step of the review methodology, the analysis of the results is carried out to complete the workflow of the methodology. This step addresses all the defined survey questions (SQs), exploring explanations and specific details. The subsequent section will present the main findings after applying the defined methodology and achieving the SLR objectives.
\section{SURVEY RESULTS}
The analysis of the survey results is based on the survey questions
\subsection{How can quantum algorithms improve the efficiency of NLP tasks?}.
The culmination of quantum computing and natural language processing generates features that act as valuable assets and provide solutions to improve the performance of NLP models.
\subsubsection{Foundational Concepts in Quantum Computing}
\mbox{} \\
\textbf{Quantum Computing principles}
\begin{itemize}
    \item \textit{Qubit:} In quantum computing, the Hilbert space\cite{nielsen2010quantum} serves as the computational universe or space. Within this space, Qubit\cite{james2001measurement} is the fundamental information unit. Unlike bit, Qubit is represented by a 0, 1 or superposition of both. From quantum mechanics principles \cite{dirac1926theory}, the Dirac notation \cite{dirac1939new} (also known as bra-ket) represents a Qubit as |0> (ket 0 notation) and |1>(ket 1 notation).
    \item \textit{Superposition a:}  Based on Schrodinger's thought experiment of cat\cite{schrodinger1935naturwissenschaften}, the superposition principle is used to represent the Qubit by combining states in a linear way
      \[
      |\psi \rangle = a|0 \rangle + b|1 \rangle
      \]
       a, b represent complex numerals. 
     \item \textit{Entanglement:} Quantum entanglement\cite{jozsa1997entanglementquantumcomputation} is an enigmatic quantum behaviour when the states of two qubits become interdependent in such a way that the state of one qubit can not be defined irrespective of the other Qubit state even if they are far apart.  

     \item \textit{Measurement:} It is a process that involves the collapse in the quantum state of the Qubit into either of the basic states, 0 or 1. In essence, the act of measurement modifies the state of the qubit, transforming it from a superposition of \( \left| 0 \right\rangle \) along with \( \left| 1 \right\rangle \) to the definite state that corresponds to the outcome of the measurement. Specifically, if \( \left| \psi \right\rangle \) is found to be in the state \( \left| 0 \right\rangle \) during the measurement, the post-measurement qubit state becomes \( \left| 0 \right\rangle \). Thus, any successive measurements executed on the same basis will generate the result '0' having probability 1.
 \end{itemize}
\textbf{Quantum Gates}

 In traditional computing, electronic gates are utilised to manipulate classical bits, which can represent either 0 or 1. Although in quantum computing, states 0 and 1 can be simultaneously placed into superposition, thereby giving rise to various interesting possibilities.
\begin{itemize}
    \item Single-qubit gates: In the domain of quantum computing, these are considered elementary gates that operate on a single qubit \cite{10.5555/1973124}\cite{barenco1995elementary}and are characterised by their representation as a 2x2 unitary matrix. Pauli gates, Hadamard gates, Rotation gates
    \item Multi-Qubit gates: These gates operate concurrently on more than one Qubit at the same time. It facilitates in creating the entanglement that connects multiple quantum states. 
    Table 2 lists the types of quantum gates, their symbols, and purpose.
\end{itemize}
Table~\ref{tab:quantum_gates} presents a comprehensive overview of both single- and multi-qubit gates, detailing their symbols, matrix representations, and respective purposes.

\begin{table}[ht]
\centering
\renewcommand{\arraystretch}{1.5}
\setlength{\tabcolsep}{6pt}
\caption{Quantum gates}
\resizebox{\textwidth}{!}{%
\begin{tabular}{|>{\centering\arraybackslash}m{2.5cm}|>{\centering\arraybackslash}m{2.5cm}|>{\centering\arraybackslash}m{2.5cm}|>{\centering\arraybackslash}m{4.5cm}|>{\arraybackslash}m{6.5cm}|}
    \hline
    \textbf{Gate Type} & \textbf{Gate} & \textbf{Symbol} & \textbf{Matrix} & \textbf{Purpose} \\
    \hline
    \multirow{6}{*}{Single Qubit} 
    & Pauli X \cite{crooks2020gates} & \includegraphics[height=0.8cm]{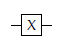} & 
    $X = \begin{bmatrix} 0 & 1 \\ 1 & 0 \end{bmatrix}$ \cite{crooks2020gates} & 
    Flips the Qubit state, similar to classical NOT gate \\
    \cline{2-5}
    & Pauli Y \cite{crooks2020gates} & \includegraphics[height=0.8cm]{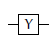} & 
    $Y = \begin{bmatrix} 0 & -i \\ i & 0 \end{bmatrix}$ \cite{crooks2020gates} & 
    Applies both a bit and phase flip to the Qubit \\
    \cline{2-5}
    & Pauli Z \cite{crooks2020gates} & \includegraphics[height=0.8cm]{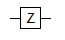} & 
    $Z = \begin{bmatrix} 1 & 0 \\ 0 & -1 \end{bmatrix}$ \cite{crooks2020gates} & 
    Applies phase flip to the Qubit \\
    \cline{2-5}
    & Hadamard (H) \cite{crooks2020gates} & \includegraphics[height=0.8cm]{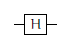} & 
    $H = \frac{1}{\sqrt{2}} \begin{bmatrix} 1 & 1 \\ 1 & -1 \end{bmatrix}$ \cite{crooks2020gates} & 
    Imposes a superposition state \\
    \cline{2-5}
    & $R_x(\theta)$ \cite{crooks2020gates} & \includegraphics[height=0.8cm]{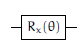} & 
    $\begin{bmatrix} \cos(\theta/2) & -i\sin(\theta/2) \\ -i\sin(\theta/2) & \cos(\theta/2) \end{bmatrix}$ \cite{crooks2020gates} & 
    Rotates the state about x-axis by $\theta$ angle \\
    \cline{2-5}
    & T Gate \cite{crooks2020gates} & \includegraphics[height=0.8cm]{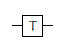} & 
    $T = \begin{bmatrix} 1 & 0 \\ 0 & e^{i\pi/4} \end{bmatrix}$ \cite{crooks2020gates} & 
    Performs a rotation of the qubit state by $\pi/4$ about the Z-axis \\
    \hline
    \multirow{3}{*}{Multi Qubit} 
    & CNOT \cite{crooks2020gates} & \includegraphics[height=0.8cm]{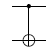} & 
    $\begin{bmatrix} 
    1 & 0 & 0 & 0 \\ 
    0 & 1 & 0 & 0 \\ 
    0 & 0 & 0 & 1 \\ 
    0 & 0 & 1 & 0 
    \end{bmatrix}$ \cite{crooks2020gates} & 
    Acts on two qubits, applying a NOT to the target only if the control Qubit is in state $\vert 1 \rangle$ \\
    \cline{2-5}
    & SWAP \cite{crooks2020gates} & \includegraphics[height=0.8cm]{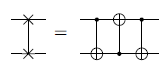} & 
    $\begin{bmatrix} 
    1 & 0 & 0 & 0 \\ 
    0 & 0 & 1 & 0 \\ 
    0 & 1 & 0 & 0 \\ 
    0 & 0 & 0 & 1 
    \end{bmatrix}$ \cite{crooks2020gates} & 
    Swaps two qubit states \\
    \cline{2-5}
    & Toffoli \cite{crooks2020gates} & \includegraphics[height=0.8cm]{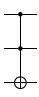} & 
    \scriptsize$
    \begin{bmatrix} 
    1 & 0 & 0 & 0 & 0 & 0 & 0 & 0 \\ 
    0 & 1 & 0 & 0 & 0 & 0 & 0 & 0 \\ 
    0 & 0 & 1 & 0 & 0 & 0 & 0 & 0 \\ 
    0 & 0 & 0 & 1 & 0 & 0 & 0 & 0 \\ 
    0 & 0 & 0 & 0 & 1 & 0 & 0 & 0 \\ 
    0 & 0 & 0 & 0 & 0 & 1 & 0 & 0 \\ 
    0 & 0 & 0 & 0 & 0 & 0 & 0 & 1 \\ 
    0 & 0 & 0 & 0 & 0 & 0 & 1 & 0 
    \end{bmatrix}$ \cite{crooks2020gates} & 
    Flips the target Qubit only if both control qubits are in $\vert 1 \rangle$ state \\
    \hline
\end{tabular}
}
\label{tab:quantum_gates}
\end{table}

\textbf{Quantum RAM (QRAM)}\\
Quantum Random Access Memory (QRAM) serves as the quantum counterpart to classical RAM\cite{giovannetti2008quantum}. By employing qubits for accessing memory units, QRAM is capable of simultaneously accessing a superposition of multiple memory units. Analogous to classical RAM, QRAM is comprised of three principal components: the input register, the memory array, and the output register. QRAM has the capacity to access numerous memory cells simultaneously in superposition for concurrent data processing, in contrast to classical RAM, which retrieves one cell at a time. This distinctive feature facilitates exponential speed enhancements in quantum algorithms, such as pattern recognition and quantum Fourier transforms, relative to classical approaches.
\subsubsection{Quantum advantage in NLP Tasks}
The capability of quantum computing technology to tackle computational issues quicker than traditional computers. This phenomena is referred as quantum advantage.
Grover's Algorithm \cite{kwiat2000grover}serves as a quintessential representation of quantum approach that offers a quadratic speedup in dealing with unstructured search challenges, making it particularly applicable to question answering task. \\
\textbf{Grover's Algorithm}
The design of Grover's algorithm \cite{kwiat2000grover} is intended for conducting searches within an unsorted database containing \(N\) elements to identify a specified item. Conventionally, a classical computer requires approximately \(\frac{N}{2}\) queries to identify the target element. In comparison, Grover's algorithm completes this process using approximately \(\sqrt{N}\) queries, thereby demonstrating a quadratic enhancement in efficiency relative to classical methodologies.
\textbf{Grover's algorithm for question answering}\\ A quantum framework integrated with Grover's algorithm\cite{correia2022groversalgorithmquestionanswering} achieves quadratic speedup in finding answers to the \textit wh question. This task involves the representation of questions by encoding them into a superposition of all potential answers.
The unique features of quantum computing, such as superposition, entanglement, and interference, play a critical role in advancing the fields of quantum computing and quantum information processing. These features provide capabilities that exceed those of traditional computing systems. Quantum Natural Language Processing (QNLP) aspires to revolutionise Natural Language Processing (NLP) by facilitating accelerated processing speeds and enhanced model performance, particularly in tasks such as sentiment analysis, text classification, and language modelling.
\subsection{How can classical text data be effectively encoded into quantum states for use in quantum machine learning models?}
Classical text data may be converted into quantum states by employing various encoding schemes to map information, such as words and sentences, into a quantum Hilbert space.\subsubsection{Text Encoded into quantum states}
Quantum encoding plays a pivotal role in building a high performance QNLP model, as it facilitates efficient representation of data.  The intersection of QNLP and information theory can be exploited to optimise tasks such as textual processing, secure data transfer, and document representation. Quantum encoding techniques for classical textual data facilitate efficient representation of data and capture complex semantic relationships. Encoding of text into quantum states can be categorised into several broad categories for classification in the context of Quantum Natural Language Processing (QNLP). These categories include methods related to quantum teleportation, quantum measurement,  quantum circuit, quantum semantic and tensor-based coding, quantum entanglement, and quantum probability. 

\begin{figure}[ht]
    \centering
  \includegraphics[width=0.75\textwidth]{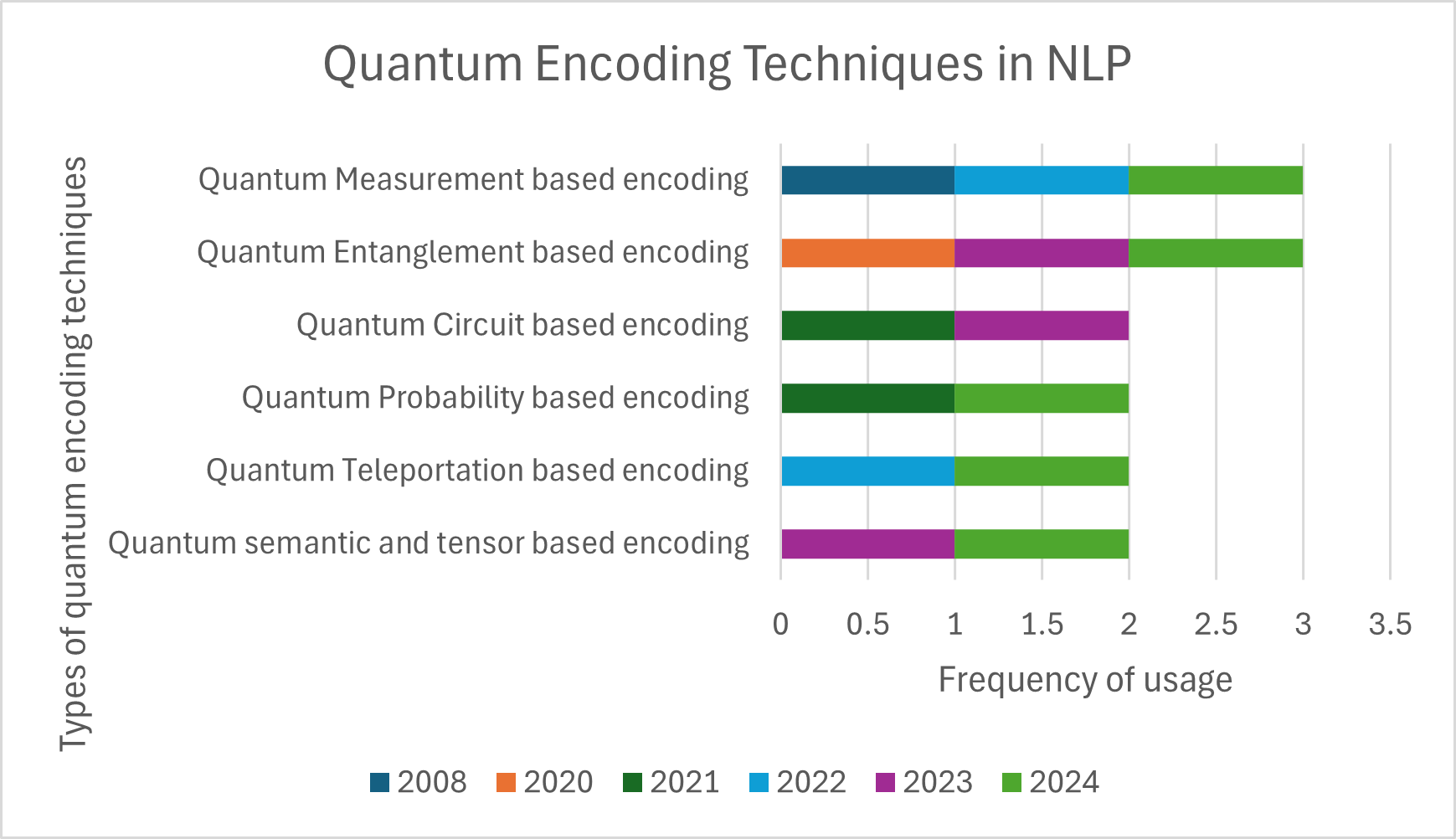}  
    \caption{Quantum Encoding}
    \label{fig:quantum_encoding1}
\end{figure}

\textit{Quantum Teleportation-Based Encoding:} The proposed Quantum Text Teleportation Protocol (QTTP) \cite{karthik2022quantum} encodes text in quantum states by utilising quantum teleportation techniques. The method entails translating the text into quantum bits (qubits), enabling secure teleportation. Incorporating Huffman Coding provides an additional layer by compressing the text data, allowing it to be transmitted more efficiently while ensuring the encoded content stays secure and can only be deciphered with the correct prefix codes. This combination supports both safe transfer and effective data management in quantum communication.\\
\textit{Quantum measurement-Based Encoding }:  A theoretical framework inspired by quantum computing \cite{huertasrosero2008characterisingerasingtheoreticalframework} models textual documents such that the lexical measurements of text are analogous to the physical measurements performed in quantum states. This framework facilitates an advanced mechanism for representing documents by leveraging quantum mechanics features to enhance the indexing and retrieval of documents. An alternative methodology \cite{correia2022quantum} employs contraction schemes aimed at integrating word representations, thereby facilitating the preservation of multiple meanings within quantum superposition. This approach leverages tensor contractions to formulate these representations, addressing syntactically ambiguous expressions. The resultant quantum states are used within quantum circuits to conduct sentence meaning disambiguation and question-resolving systems, thereby augmenting the effectiveness of natural language processing.
Quantum encoding based on superposing technique \cite{kim2024quantumsuperposingalgorithmquantum} encodes classical data by generating superpositions of Qubit indices. This paradigm obviates the need for direct access to the actual data through the quantum computation phase and allows for seamless incorporation of classical and quantum functionalities. 
The paper discusses quantum encoding, which efficiently maps classical data, such as text, into quantum states. It emphasises a quantum superposing algorithm that enhances computational efficiency in encoding finite-size databases into Qubit registers, optimising the process significantly.
\\
\textit{Quantum Circuit-Based Encoding}:  A method for encoding vectors utilising hardware effective variational quantum circuits \cite{melnikov2023quantum} and are tuned by employing Riemannian optimisation with autogradient calculation. Further, the barren plateau problem is overcome by using the 'cut once, measure twice' method. Another technique \cite{Laakkonen_2024} encodes word embeddings as parameterised quantum circuits. Within this encoding framework, textual information is characterised by quantum states, thereby facilitating compositionality in accordance with the linguistic structure of the text. Another DisCoCat model-based encoding technique \cite{du2022gentleintroductionquantumnatural} encodes textual information into quantum states by representing sentence meanings as vectors within quantum computational systems. This approach extends the distributional semantics inherent to words by integrating them into the compositional semantics of sentences, utilising tensor products to combine word vectors in accordance with the sentence's syntactic structure. Despite this encoding algorithm's limited efficiency on classical computational devices, it demonstrates substantial scalability when implemented on quantum circuits, thereby facilitating the effective representation and processing of natural language within a quantum framework. A hybrid workflow method \cite{o2020hybrid} performs encoding by representing corpus meanings using quantum circuit models. This entails the processing and decoding of data sets that encompass both limited and extensive scales, thereby facilitating efficient comparative analyses among sentences with a specified structure.\\
An amplitude encoding methodology \cite{9996813} for text classification endeavors integrates an amplitude-encoded feature map in conjunction with a quantum support vector machine, thereby enhancing accuracy in sentiment prediction within a movie reviews dataset. 
An alternative encoding method \cite{ghosh2021encodingclassicaldataquantum} facilitates the transposition of classical data into a quantum computational framework by regarding a set of N classical data, articulated as a column matrix, to produce an n-bit quantum state. This method uses Schmidt decomposition and singular value decomposition techniques to construct a family of quantum circuits.
\textit{Quantum Semantic and Tensor Based Encoding}: A combination of tensor operations with the semantic distribution classification model \cite{han2023quantum} for Quantum Natural Language Processing (QNLP), which allows efficient coding of semantic spaces for the effective extraction of text structures.
\\
\textit{Quantum Entanglement based encoding }: An Entanglement Embedding (EE) technique \cite{Chen_2023} captures non-classical correlations among words to transform sequences of words into entangled pure states.
Another technique \cite{yu2020quantum} incorporating entanglement of adjacent words improves the dimensionality of sentence representations facilitating enhanced representation of semantic word relations. The method performs conversion by effectively converting textual data into quantum states by incorporating dimensionality reduction techniques to reduce semantic noise. 
\\
\textit{Quantum Probability-Based Encoding}: The technique based on the quantum probability principle\cite{yan2021quantum}, encodes text in quantum states. By capturing complex relations and semantics in the data, textual information is encoded as quantum states.  This novel encoding methodology seeks to enhance the efficiency of document classification by leveraging the distinctive attributes inherent to quantum states. Table~\ref{tab:quantum_encoding} illustrates the quantum encoding techniques and their corresponding year. Figure Figure~\ref{fig:quantum_encoding1} presents an analysis of quantum encoding techniques employed within natural language processing, along with their frequency of application over the years.

\begin{table}[ht]
    \caption{Quantum Encoding Methods and References}
    \centering
    \begin{tabular}{|c|c|c|}
        \hline
        \textbf{Year} & \textbf{Encoding} & \textbf{Reference} \\
        \hline
        2022 & Quantum Teleportation based encoding & \cite{karthik2022quantum} \\
        \hline
        2008, 2022, 2024 & Quantum Measurement based encoding & \cite{huertasrosero2008characterisingerasingtheoreticalframework}, \cite{correia2022quantum}, \cite{kim2024quantumsuperposingalgorithmquantum} \\
        \hline
        2021, 2023, 2024 & Quantum Circuit based encoding & \cite{melnikov2023quantum}, \cite{Laakkonen_2024}, \cite{ghosh2021encodingclassicaldataquantum} \\
        \hline
        2023 & Quantum Semantic and Tensor based encoding & \cite{han2023quantum} \\
        \hline
        2023, 2020 & Quantum Entanglement based encoding & \cite{Chen_2023}, \cite{yu2020quantum} \\
        \hline
        2021 & Quantum Probability based encoding & \cite{yan2021quantum} \\
        \hline
    \end{tabular}
    \label{tab:quantum_encoding}
\end{table}
 \subsubsection{Classical to quantum information encoding}
 The methodologies for Quantum encoding techniques of non-textual classical data/information is tailored to the specific type of input data and also to the intended application. 
 Quantum encoding for dynamic directed graphs\cite{della2024quantum} represents directed graphs as unitary matrices to enable efficient manipulation and traversal in quantum computing frameworks. Although the scalability of these quantum algorithms for larger and complex graphs is not explored, which could be a significant research gap.

Optimisation algorithm using tensor network techniques \cite{ben2024approximate} focuses on employing local gates in the preparation of target states, also addressing barren plateau in quantum optimisation. A limitation of this approach is it does not address the scalability to larger systems beyond one dimensional matrix product states, thus impacting its applicability.

Hybrid Quantum Encoding\cite{bhabhatsatam2023hybrid}  combines amplitude and basis encoding to represent indices for data values. This combination leads to enhanced data storage and processing. Further,  research is needed to enhance encoding and decoding processes to enhance efficiency and effectiveness in information processing tasks. 
Exponential data encoding\cite{shin2023exponential} use small number of encoding gates to map training data allowing efficient representation of functions. This encoding strategy provides a training resource advantage over classical methods by reducing computational resources. Further, exploration is needed in scenarios where input data increases and in identifying applicability of these strategies in diverse fields. 

Another encoding technique in Parameterised Quantum Circuits(PQC)\cite{li2022concentration} highlights that at an exponential rate with respect to circuit’s depth the average encoded state can concentrate on maximally mixed state. A gap identified in this study pertains to influence of data encoding on quantum neural networks. 

Dark web content classification using quantum encoding \cite{dalvi2023dark} utilises the Universal Sentence Encoder to convert the crawled dark web data into quantum data. A quantum circuit is fed with this quantum representation, which uses a soft max function to perform comparison of actual category labels with the output obtained. This approach offers enhanced efficiency in time and memory consumption. This study identifies limitation of the approach for comprehensive and diverse keyword datasets.

Quantum encoding for vectorise data \cite{balewski2024quantum}, uses two methods Q Crank and QBArt to perform encoding as rotation on data qubits and for binary representation thus allowing fewer measurements and efficient arithmetic operations. Although, there is a limitation on the scalability and practicality of this approach.

Quantum encoding translates quantum programs from the Proto-Quipper-D language into SZX-diagrams\cite{Borgna_2023}, a graphical representation used in the Scalable ZX-calculus. Another optimal quantum encoding \cite{farokhi2024optimaluniversalquantumencoding} generates quantum states, crucial for quantum based machine learning applications. This strategy maximises the maximal quantum leakage, which determines encoding quality measure. 

Quantum encoding for the transformation of molecular docking problems into quadratic unconstrained binary optimisation (QUBO) model \cite{zha2023encoding}plays a pivotal role in Tackling Coherent Ising Machine (CIM) issue.  These methods enhance pose sampling within the realm of molecular docking, significantly improving speed and efficiency in comparison to classical computing methods. Quantum angle encoding\cite{ovalle2023quantum}in the context of quantum-classical convolutional neural network, performs encoding thereby enabling proficient representation and modulation of information. 
Table ~\ref{table:quantum-encodings} illustrates quantum-based encoding corresponding to various input types, along with their respective applicability.

From dynamic directed graphs to complex data, these techniques offer revolutionary solutions for wide ranging applications as depicted in Table ~\ref{table:quantum-encodings}
\renewcommand{\arraystretch}{1.3} 
\begin{table}[ht]
\caption{Quantum Encoding Methods and Their Applications} \label{table:quantum-encodings} \resizebox{\textwidth}{!}{ \begin{tabular}{|p{2cm}|p{2cm}|p{4cm}|p{6cm}|p{5cm}|} \hline \textbf{Reference} & \textbf{Year} & \textbf{Type} & \textbf{Input Type} & \textbf{Applicability} \\ \hline \cite{della2024quantum} & 2023 & Graph Representation Encoding & Directed graphs (e.g., adjacency matrices, edge lists, or node/edge attributes) & Used to depict graphs as quantum states, allowing efficient traversal, multi graph encoding, and quantum random walks. \\ \hline \cite{ben2024approximate} & 2024 & State Preparation Encoding & Classical data or quantum states that need to be prepared for quantum algorithms & Prepares quantum states on near-term quantum devices. \\ \hline \cite{bhabhatsatam2023hybrid} & 2023 & Hybrid Encoding (Amplitude and Basis Encoding) & Classical data (such as numerical values, vectors, or matrices) & Generates quantum states for quantum information processing and machine learning. \\ \hline \cite{shin2023exponential} & 2023 & Exponential Encoding in Quantum Supervised Learning & Training data (e.g., vectors or datasets in machine learning) & Maps training data into quantum states for quantum supervised learning tasks. \\ \hline \cite{li2022concentration} & 2022 & Parameterised Quantum Circuits (PQC) for Data Encoding & Classical data in the form of numerical vectors or matrices & Encodes using quantum circuits. \\ \hline \cite{dalvi2023dark} & 2023 & Dark Web Classification Encoding & Crawled data from the dark web (e.g., text data, URLs, or metadata) & Classifies dark web data (Tor hidden services) into predefined categories using quantum classifiers. \\ \hline \cite{balewski2024quantum} & 2024 & QCrank and QBArt Encoding & Real-valued data (for QCrank) or binary data (for QBArt) & DNA pattern matching, binary image retrieval, and other data-intensive tasks. \\ \hline \cite{Borgna_2023} & 2022 & Quantum Program Encoding in ZX-calculus & Quantum programs or algorithms (written in languages like Proto-Quipper-D) & Encodes quantum programs into SZX-diagrams for efficient quantum computation and algorithm design. \\ \hline \cite{farokhi2024optimaluniversalquantumencoding} & 2024 & Quantum Encoding for Statistical Inference & Classical data, often in the form of datasets used for statistical analysis or machine learning & Transforms classical data into quantum states to enhance statistical inference accuracy. \\ \hline \cite{zha2023encoding} & 2023 & Molecular Docking Encoding (QUBO) & Molecular data (e.g., 3 D coordinates of molecules, atom types, or protein-ligand interaction data) & Encodes molecular docking problems into QUBO models for optimisation on quantum computers. \\ \hline \cite{ovalle2023quantum} & 2023 & Quantum Angle Encoding in Neural Networks & Classical data, represented as vectors or matrices, to be processed by quantum neural networks & Encodes data into quantum states in the form of angles for quantum-classical neural networks. \\ \hline \end{tabular} } \end{table}
\subsection{What are the general categories of Quantum Natural Language Processing(QNLP) models }
Quantum Natural Language Processing (QNLP) models can be categorised based on their principles, architecture, and computational approaches. 
\subsubsection{Categorical QNLP Models}
The compact closed structure in Hilbert space, represented through bipartite projectors\cite{mackey1957quantum}, thereby facilitates the structural characterisation of information flows in entangled quantum systems\cite{abramsky2003physical}. A comprehensive analysis\cite{coecke2014logic} of various quantum protocols was conducted, utilising graphical notation to depict the information flow paths through a network of projectors. A significant advancement in the development of the categorical approach\cite{abramsky2004categorical} and the strongly compact closed category\cite{abramsky2009abstractphysicaltraces} prompted more profound investigations into the application of category theory to quantum mechanics\cite{abramsky2009categorical}, resulting in a novel perspective on the structural aspects of quantum information and computation. The quantum framework for natural language\cite{coecke2010mathematical} effectively combines linguistic meanings with grammatical structures by utilising mathematical concepts such as compact closed categories, tensor products, and diagrammatic calculus. This integration lays the groundwork for Quantum Natural Language Processing (QNLP), wherein the compositional structure of language is encapsulated within the mathematical framework.
\begin{enumerate}
    \item The categorical compositional distributional semantics framework (DisCoCat) \cite{Coecke_2018} serves as a rigorous formalism for the analysis of the compositional structure inherent in natural language \cite{coecke2010mathematical}. It has demonstrated substantial success across multiple domains within natural language processing. Within the DisCoCat formalism, the grammar of natural language is almost always mathematically formalised by using Lambek's pregroup grammars\cite{10.5555/645666.665029}.\\
 \item \textit{Compositional Distributional Circuits (DisCoCirc)} Model:
A radical progression from the DisCoCat framework led to a mathematical foundation named Compositional Distributional Circuits (DisCoCirc)\cite{coecke2020mathematicstextstructure}. A primary goal is to resist perceiving word meanings as fixed entities, as typically practised in DisCoCat, and instead to conceptualise them as fluid entities that adjust and change in reaction to the contextual information provided by the text about the word. This approach enables a more flexible and context-sensitive representation of meaning, which captures the nuances of language use more effectively. A simple translation from English to Urdu grammar uses DisCoCirc framework\cite{Waseem_2022} to eliminate grammatical differences between languages.
\end{enumerate}
\subsubsection{Quantum Probabilistic Model}: 
Quantum Probabilistic model has an advantage in terms of their geometric representation which provides an ability to move from one basis to another. Unlike classical probability theory, this ability provides great flexibility in decision making, as mentioned in \cite{moreira2016quantum}
\begin{enumerate}
 \item \textit{Quantum Boltzmann Machine(QBM):}
    A Quantum Boltzmann Machine (QBM)\cite{amin2018quantum} is a quantum probabilistic model for machine learning that relies on the Boltzmann distribution derived from a quantum Hamiltonian. Using the quantum characteristics of the processor during the modelling and training stages. The QBM training approach, detailed in \cite{kieferova2017tomography}, uses the relative entropy between the data density matrices and those of the quantum system as its objective function.\item \textit{Fock Space Representation}: Computational linguistics problems can be best mapped onto quantum computers using the fock space representation\cite{wiebe2019quantumlanguageprocessing}, which provides efficient encoding with fewer qubits. Further, this representation led to development of a formalism for harmonic grammars and introduces a training technique for quantum Boltzmann machines which may have applications beyond language processing.

\end{enumerate}
\subsubsection{Quantum Circuit Models:}
The quantum circuit model serves as an essential framework in the domain of quantum computing, offering a structured depiction of quantum algorithms and processes. Compared to classical Boolean circuits, this model operates on qubits, the fundamental units of quantum information. It plays a pivotal role in quantum computation, communication, and cryptography by providing a machine code-level representation of quantum algorithms and protocols. The adaptability of the model enables it to simulate alternative quantum computing paradigms, such as adiabatic and measurement-based models, and functions as a foundational blueprint for the advancement of quantum hardware\cite{CircuitModelofQuantumComputation}.
\begin{enumerate}
    \item \textit{Variational Quantum Algorithms(VQA): }
    One of the opportunistic approaches to achieve a near-term quantum advantage in Noisy Intermediate-Scale Quantum (NISQ) devices is the Variational Quantum Algorithm\cite{cerezo2021variational}. It has a unified structure to represent a task as a parameterised cost function to be evaluated on a quantum computer, while a classical optimiser trains the VQA parameters. It has the potential to offer a wide range of applications in different domains.
   \item \textit{Quantum Neural Networks(QNNs):}
    Quantum neural computation\cite{kak1995quantum} inaugurated an innovative paradigm through the integration of neural networks with quantum computation, thereby revealing new opportunities for scholarly inquiry. Subsequently, \cite{menneer1995quantum} proposed quantum-inspired methodologies for training neural networks that have the potential to increase computational efficiency and tackle challenges typically beyond the scope of conventional neural network frameworks. Quantum neural networks\cite{Abbas_2021}, identified as a variant of variational quantum algorithms, utilise quantum circuits comprising parameterised gates. Initially, data is encoded into a quantum state by means of a feature map, which serves to augment performance. Subsequently, a variational model with parameterised gates is applied and optimised for specific tasks through loss function minimisation. The final output is obtained by applying classical post-processing to the measurement results.
\end{enumerate}
\subsubsection{Quantum Kernel Methods} Quantum kernel methods represent a promising paradigm in the field of quantum machine learning, capitalising on the capabilities of quantum computing to improve the efficacy of classical kernel techniques. These methodologies facilitate the mapping of data into high-dimensional quantum feature spaces, wherein quantum devices estimate the kernel function, while the remainder of the computations are executed using classical resources. The potential for a quantum advantage is manifested when the quantum kernel function eludes efficient estimation by classical methodologies. This advantage is notably prominent in scenarios where quantum processing is able to discern patterns that manifest as noise in classical computational systems.
\begin{enumerate}
\item \textit{Quantum Support Vector Machines(QSVM):}
The QSVM technique\cite{anguita2003quantum} innovatively utilises quantum computing principles to enhance SVM training, offering potential advantages in terms of efficiency and effectiveness, particularly for complex and digital implementations. The support vector machine (SVM) is trained through the digital representation of the problem, utilising quantum optimisation techniques to effectively navigate the search space of potential solutions.
Another new quantum algorithm discussed in \cite{rebentrost2014quantum} shows that the support vector machine, an optimised binary classifier, can be implemented on the quantum computer. This formulation is popularly known as the Quantum Support Vector Machine(QSVM).
The quantum support vector machine (SVM) methodology involves the reformation of the SVM into a least-squares problem, enabling efficient matrix inversion via quantum computing. This approach significantly accelerates both the training and classification stages, particularly for big data applications. It uses a non-sparse matrix exponentiation technique to manage the kernel matrix and employs a low-rank approximation through principal component analysis(PCA) to proficiently classify data within a higher-dimensional space.
\end{enumerate}
\subsubsection{Quantum Language Model} Quantum language models take advantage of the core principles of quantum mechanics, such as quantum probability, superposition, and entanglement, to improve the understanding and processing of linguistic constructs. By employing density matrices, QLMs represent words as quantum states, which aids in the modelling of semantic relationships among sentences. This method aligns with measurement operators found in quantum theory, providing a semantic framework for interpreting interactions between words\cite{10.1007/978-3-031-30111-7_4}\cite{Zhang_Niu_Su_Wang_Ma_Song_2018}.
\begin{enumerate}
\item
\textit{Quantum-inspired language model: }
A quantum mechanics-inspired language model and probability calculus are introduced in \cite{basile2017towards}. This research offers a conceptual framework to assess the utility of using orthogonal projectors in language modelling. The approach is kept straightforward by linking each word to a separate basis vector, with the Hilbert space's dimensionality being aligned with the vocabulary size. A generalised language modelling strategy \cite{10.1145/2484028.2484098} presents an innovative use of quantum probability in information retrieval, where dependencies are treated as superposition events, helping to integrate them during estimation. This paper is noted as the inaugural practical application of quantum probability to IR, showcasing significant progress over both a robust bag-of-words baseline and a stronger non-bag-of-words model. 

\item \textit{Density Matrix: } A depiction of the density matrix as a versatile tool is found in \cite{10.1007/978-3-642-54943-4_13}, combining the strengths of vector space and language model representations, thus heralding a new era of retrieval models. A Session-based Quantum Language Model (SQLM) \cite{10.1145/2766462.2767819} addresses the multi-query session search task by utilising a density matrix transformation model to depict the evolution of a user's information needs based on search engine interaction. This model draws on features from both positive and negative feedback. 
\end{enumerate}
\subsubsection{Hybrid Classical Quantum Computing Model} NISQ Era Hybrid Quantum-Classical Machine Learning Research \cite{de2022survey} highlights the importance of hybrid quantum-classical systems in the current quantum computing framework, presenting a means to optimise the capabilities of the quantum system while managing the constraints of the NISQ era. Hybrid systems capitalise on the strengths of both quantum and classical computing. Parameterised quantum circuits (PQCs) serve as bridges, allowing classical algorithms to steer quantum operations and facilitating the training of quantum algorithms using classical techniques. A hybrid classical-quantum strategy for natural language processing \cite{o2020hybrid} illustrates the use of quantum computing to execute NLP tasks, symbolising corpus meanings and evaluating sentence structures. This work describes the design of a hybrid workflow for representing corpus data, both small and large-scale, for encoding, processing, and decoding via a quantum circuit model. Additionally, the study demonstrates the efficiency of the results and provides the developed toolkit as open source software.
\subsection{What are the different NLP tasks handled using quantum computing?}
Quantum computing has been used in a spectrum of natural language processing (NLP) tasks through various classifications of quantum NLP (QNLP) models, each utilising distinct quantum characteristics. The breadth of applications illustrates how particular model types correspond to the specific computational requirements inherent in various NLP tasks.
\subsubsection{Sentiment Analysis}
Quantum computing offers promising advancements in sentiment analysis, a key application of natural language processing (NLP). Using the distinctive capabilities of quantum computing, researchers endeavour to improve the efficiency and precision of sentiment analysis tasks. This involves integrating quantum computing with classical machine learning techniques, exploring quantum-inspired models, and developing novel quantum representations for sentiment classification. The following sections delve into these approaches and their implications.\\
\textbf{Categorical QNLP Models}
    \begin{itemize}
        \item{Distributional Compositional Categorical (DisCoCat) Model} \\Another quantum computing methodology \cite{9951286} employs quantum operations, such as the X gate and the Z-X gate, to represent sentiment words within quantum circuits, there by enhancing the accuracy of sentiment analysis there. The research explores four distinct approaches, with the third approach, which involves appending an X-gate at the conclusion of the sentence, demonstrating superior performance. By training state vectors using classical classifiers such as SVM and SPSA, the study achieves accuracy levels of up to 81.67\% were achieved. This method showcases how quantum representation in sentiment analysis can lead to improved accuracy, offering a promising avenue for sentiment analysis applications on quantum computers.
        \item The Simple Sentiment Analysis Ansatz for sentiment classification \cite{ruskanda2023simple}is a novel and less complex quantum representation using Variational Quantum Algorithms, reducing parameters and gates compared to other approaches.
        \end{itemize}
\textbf{Quantum Probabilistic Model} 
\begin{itemize} 
\item A multimodal analysis based on the quantum probability framework of sentiment, sarcasm, and emotion\cite{liu2023quantum} proposes a multitask learning environment based on quantum probability for integrated multimodal sentiment, sarcasm and emotion analysis, with the goal of overcoming the challenges of multimodal affect understanding.
\end{itemize}
\textbf{Quantum Circuit Models}
\begin{itemize}
    \item Variational Quantum Algorithms (VQA) 
    \begin{itemize}
        \item Variational quantum classifiers (VQC)\cite{Joshi_2021} along with parameterised circuits like EfficientSU2 and RealAmplitudes. The QML models demonstrated superior performance in sentiment analysis tasks in comparison to other models such as Support Vector Machines, Gradient Boosting, and Random Forests. Variational quantum classifiers (VQC) exploit quantum computational advantages to efficiently process complex functions, thereby achieving greater accuracy. When these quantum models are trained on quantum devices, they are capable of optimising parameters and minimising loss, leading to enhanced performance in sentiment analysis. In particular, within the study, the EfficientSU2 model, trained in more than 100 epochs, achieved the highest accuracy of 74. 
        \item A quantum inspired method \cite{singh2022emotion} has been developed for the quantification of emotional intensities in real-time through the application of quantum state fidelity estimation. Using quantum computing methodologies, notably the Quantum Variational Classifier, it has been possible to successfully quantify emotional intensities such as joy, sadness, contempt, anger, surprise, and fear. This methodology demonstrates the capacity of quantum computing to augment the precision of sentiment analysis. This is achieved by providing a more thorough understanding of emotional nuances and their varying intensities, surpassing the limitations of traditional categorical recognition models.
    \end{itemize}
\item Quantum neural networks (QNNs)
   \begin{itemize}
       \item A Duplication free Quantum based Long Short-Term Memory (DQLSTM) \cite{hou2022realization}for sentiment analysis reduces the requirement in the number of qubits compared to conventional methods,  thus improving efficiency. The methodology utilises amplitude encoding to represent classical data, thereby enhancing the precision in applications such as Chinese sentiment analysis. Hence, quantum computing, when applied through techniques such as DQLSTM, has the capacity to enhance sentiment analysis accuracy by optimising the utilisation of qubits, thereby achieving a level of performance comparable to that of classical models.
       \item The Complex-valued Quantum enhanced Long Short-term Memory (CQLSTM) model \cite{chu2024effective} which employs complex-valued embeddings and quantum-inspired techniques, proficiently encodes semantic dependencies and captures interactions between words and sentences. This methodology augments the accuracy of sentiment analysis by incorporating additional semantic information, capturing contextual semantic interactions, and encoding long-term dependencies within emotional features. Consequently, as evidenced by the CQLSTM model, quantum computing can enhance sentiment analysis accuracy through the effective application of complex-valued embeddings and quantum-inspired neural network architectures.
       \item The Quantum based Fuzzy Neural Network (QFNN) \cite{tiwari2024quantum} integrates QNN with fuzzy logic. By employing complex numbers within the Fuzzifier to discern sentiment and sarcasm features, and utilising QNN in the Defuzzifier for making predictions, the QFNN algorithm demonstrates superior performance compared to several contemporary methods in tasks related to sentiment and sarcasm detection. Additionally, the QFNN exhibits notable robustness in environments with noise, characterised by its expressible and entanglement capabilities, thereby enhancing the accuracy of sentiment analysis through the application of quantum computing techniques.
       \item Complex-Valued Neural Networks (Complex QNN)\cite{lai2023quantum} are capable of utilising the imaginary component of complex numbers to encapsulate latent information, thereby enabling more sophisticated data representation.Through the mapping of words into complex vector spaces, quantum-inspired models such as ComplexQNN facilitate the integration of quantum computing theory with natural language processing. This methodology has been shown to enhance sentiment analysis outcomes, as evidenced by findings in 10\% accuracy increase compared to traditional models like TextCNN and GRU, and competitive performance with advanced models like ELMo, BERT, and RoBERTa.
       \item Quantum-enhanced Transformer based model \cite{di2022dawn} replaces linear transformations with Variational Quantum Circuits (VQCs), leading to a more efficient and powerful way in analysing text sentiments. This technique enables the execution of more intricate computational processes and potentially enhances the comprehension of semantic information within sentences.
       \item The Quantum-Like multi modal network framework\cite{zhang2020quantum} incorporating quantum features such as entanglement and superposition to model interaction dynamics in multiparty conversational sentiment analysis. This framework is capable of capturing intricate relationships and interactions in sentiment analysis with greater precision compared to traditional methodologies. The concurrent data processing capabilities of quantum computing, along with its potential to explore multiple possibilities simultaneously, contribute to an augmented accuracy in sentiment analysis by furnishing a more nuanced comprehension of the dynamics inherent in multiparty conversations.
    \item Quantum Inspired Multi Modal Fusion \cite{li2021quantum} models intra-modal interactions between words using quantum superposition and inter-modal interactions across different modalities through quantum entanglement. By integrating complex-valued neural networks derived from the principles of quantum theory, the model demonstrates performance that is on par with contemporary state-of-the-art systems across established benchmarking datasets. This methodology facilitates an enhanced comprehension of intricate interactions and correlations inherent in multimodal data, thereby resulting in superior sentiment predictions predicated upon sentence representations.
    \item An innovative attention mechanism, designated as DMATT\cite{8995180}, draws inspiration from Quantum Mechanics and enhances the precision of conversational sentiment analysis within deep learning models, such as DMATT-BiGRU. Notable improvements in sentiment analysis outcomes have been documented when contrasted with conventional attention mechanisms. This approach, influenced by quantum principles, utilises the density matrix concept to more adeptly capture subtle sentiment variations in conversational data. Consequently, quantum-inspired methodologies, exemplified by DMATT, have the potential to substantially improve sentiment analysis accuracy by offering more efficacious attention mechanisms in deep learning models.
    \item A Quantum Cognitively Motivated Decision Fusion approach for Video Sentiment Analysis\cite{gkoumas2021quantumcognitivelymotivateddecision} conceptualises sentiment judgements as quantum superposition states and perceives uni-modal classifiers as incompatible observables. By employing this quantum-inspired fusion strategy, cognitive biases are addressed, and the accuracy of sentiment analysis is enhanced through the effective management of all possible combination patterns, even in scenarios where individual classifiers are inadequate. Consequently, quantum computing augments sentiment analysis accuracy by introducing a pioneering methodology that encapsulates the intricacies of decision fusion in a quantum-like framework, thereby surpassing both traditional methods and the latest fusion techniques.
    \item A Quantum Multimodal Neural Network Model for Sentiment Analysis in Quantum Circuits \cite{10778283} introduces an innovative QMNN framework grounded on Parameterized Quantum Circuits (PQCs) specifically tailored for multimodal sentiment analysis applications. The architecture comprises four fundamental components, beginning with the multimodal data preprocessing module, which converts images and textual data from various modalities into vectorized representations.
\end{itemize}
    \end{itemize}
\textbf {Quantum Kernel Methods}
\begin{itemize}
    \item Quantum Support Vector Machines(QSVM):
    \begin{itemize}
    \item An amplitude-coded feature map \cite{9996813} in conjunction with a support vector machine can enhance sentiment analysis accuracy. This approach demonstrates the capability of quantum computing as a significant advancement in quantum natural language processing in comparison with past results.
    \item   The quantum machine learning model \cite{omar2023quantum} demonstrates an improved capability in predicting sentiments of Arabic-language tweets with improved accuracy. The results indicate that QC slightly outperforms traditional machine learning approaches. Although conventional machine learning techniques may provide faster processing for smaller datasets, QC appears promising in achieving higher accuracy in situations where conventional methods may face challenges. This research highlights the efficiency of QC in increasing the accuracy of sentiment analysis, particularly when processing large data sets in the Arabic language.
    \item Quantum computing model\cite{sharma2023role} that depicts the role of quantum-based entanglement in improving the efficacy of quantum kernel used for classification tasks. This technique also improves processing power and computational capabilities, leading to more precise sentiment analysis. 
    \end{itemize}
\end{itemize}

\textbf{Quantum Language Models}
\begin{itemize}
\item Quantum Entanglement
Hilbert Space representation for aspect-based sentiment analysis (ABSA) models:
The representation of words as complex-valued vectors within a Hilbert Space for aspect-based sentiment analysis (ABSA) models, as discussed by \cite{zhao2022quantum}, captures more nuanced semantic information than conventional real-valued models. These complex-valued embeddings provide additional semantic insights beyond those offered by real embeddings, thereby enhancing the performance of sentiment classification.
\item Quantum Projection and Quantum Entanglement Enhanced Network model \cite{Zhao_Wan_Qi_2024} using quantum entanglement and quantum projection techniques to project aspects onto a complex Hilbert space. This technique enhances sentiment analysis accuracy by correlating multiple language systems through quantum entanglement, leading to state-of-the-art performance in sentiment analysis tasks across multiple languages.
\item The study entitled "Unsupervised Sentiment Analysis of Twitter Posts Using Density Matrix Representation"\cite{10.1007/978-3-319-76941-7_24} deals primarily with the artificial construction of two sentiment dictionaries. It proceeds by generating density matrices for both the documents and dictionaries through an extended quantum language model (QLM). Following this, it applies quantum relative entropy to assess the similarity between the density matrices of the documents and those of the dictionaries.
\end{itemize}
\textbf{Density Matrices: }
    \begin{itemize}
        \item The Fused Sememe Knowledge Quantum-like Chinese Sentiment Analysis \cite{wang2023quantum} employs density matrix from quantum theory alongside sememe knowledge to enhance the precision of sentiment analysis. By integrating sememe, the smallest unit of semantic knowledge, into the text vector, a holistic knowledge system is developed, thereby enhancing the quality of text representation. This method augments the model's efficacy in Chinese implicit sentiment analysis by capturing nuanced semantic information, thereby demonstrating the promise of quantum-inspired methodologies in the realm of sentiment analysis.
        \item Quantum-Inspired Sentiment Representation (QSR) model\cite{zhang2019quantum} uses density matrices acquired from probability theory improves sentiment analysis by incorporating sentiment information into document representations. 
This quantum inspired technique captures the sentiment through adjectives and adverbs to effectively model semantic and sentiment information. This technique surpasses contemporary models in sentiment analysis applied to Twitter datasets, thereby illustrating the potential efficacy of quantum methodologies.
\item 
The Quantum-Inspired Multi-Modal Sentiment Analysis Framework\cite{zhang2018quantum} demonstrates a proficient capability to process and analyse multimodal data, including textual and visual input, to extract sentiments with heightened precision. The potential of quantum computing to manage complex calculations and examine a multitude of possibilities concurrently facilitates the development of advanced sentiment analysis methodologies, thereby enhancing the accuracy of understanding and interpreting emotions expressed across different data modalities.
\end{itemize}

\begin{table}[ht]
\caption{Summary of QNLP Models and Methods used in Sentiment Analysis Task}
\label{tab:SA table}
    \centering
    \resizebox{0.8\textwidth}{!}{  
        \begin{tabular}{|c|c|c|c|}
            \hline
            \textbf{Work} & \textbf{Year} & \textbf{QNLP Model} & \textbf{Methods Used} \\
            \hline
            \cite{9951286} & 2022 & Categorical QNLP Model & DisCoCat model \\
            \hline
            \cite{ruskanda2023simple} & 2023 & Categorical QNLP Model & Ansatz-based DisCoCat \\
            \hline
            \cite{liu2023quantum} & 2023 & Quantum Probabilistic Model & Quantum similarity for social bot classification \\
            \hline
            \cite{Joshi_2021} & 2021 & Quantum Circuit Model & Variational Quantum Algorithm (VQA), Random Forest, SVM \\
            \hline
            \cite{singh2022emotion} & 2022 & Quantum Circuit Model & Quantum Variational Classifier (VQC) \\
            \hline
            \cite{hou2022realization} & 2022 & Quantum Circuit Model & Deep Quantum LSTM (DQLSTM) \\
            \hline
            \cite{chu2024effective} & 2024 & Quantum Circuit Model & Complex Quantum LSTM (CQLSTM) \\
            \hline
            \cite{tiwari2024quantum} & 2024 & Quantum Circuit Model & Quantum Feedforward Neural Network (QFNN) \\
            \hline
            \cite{lai2023quantum} & 2023 & Quantum Circuit Model & Complex-valued QNN \\
            \hline
            \cite{di2022dawn} & 2022 & Quantum Circuit Model & Quantum-enhanced Transformer \\
            \hline
            \cite{zhang2020quantum} & 2020 & Quantum Circuit Model & Multimodal sentiment analysis using quantum measurements \\
            \hline
            \cite{li2021quantum} & 2021 & Quantum Circuit Model & Complex-valued Quantum Neural Network (QNN) \\
            \hline
            \cite{8995180} & 2019 & Quantum Circuit Model & DMATT (Dual Memory Attention Tensor Transformer) \\
            \hline
            \cite{gkoumas2021quantumcognitivelymotivateddecision} & 2021 & Quantum Circuit Model & Multimodal sentiment classification framework \\
            \hline
            \cite{10778283} & 2024 & Quantum Circuit Model & Quantum Neural Network (QNN) \\
            \hline
            \cite{9996813} & 2022 & Quantum Kernel Methods & Quantum Support Vector Machine (QSVM) \\
            \hline
            \cite{omar2023quantum} & 2023 & Quantum Kernel Methods & Quantum Support Vector Machine (QSVM) \\
            \hline
            \cite{sharma2023role} & 2023 & Quantum Kernel Methods & Quantum entanglement-based feature mapping \\
            \hline
            \cite{zhao2022quantum} & 2022 & Quantum Language Model & Aspect-Based Sentiment Analysis (ABSA) with QNLP \\
            \hline
            \cite{Zhao_Wan_Qi_2024} & 2024 & Quantum Language Model & Quantum Projection and Entanglement for representation \\
            \hline
            \cite{wang2023quantum} & 2023 & Quantum Language Model & Quantum representation using density matrices \\
            \hline
            \cite{zhang2018quantum} & 2019 & Quantum Language Model & Semantic modeling with density matrices \\
            \hline
            \cite{zhang2018quantum} & 2018 & Quantum Language Model & Density matrix-based language modelling \\
            \hline
            \cite{10.1007/978-3-319-76941-7_24} & 2018 & Quantum Language Model & Unsupervised quantum model using density matrices \\
            \hline
            \cite{10459858} & 2023 & Hybrid Classical Quantum Model & Quantum kernel methods with PCA \\
            \hline
            \cite{al2023quantum} & 2023 & Hybrid Classical Quantum Model & Quantum Particle Swarm Optimisation \\
            \hline
            \cite{liu2023quantum} & 2023 & Hybrid Classical Quantum Model & Semantic matching using quantum social bot architecture \\
            \hline
            \cite{bar2024quantum} & 2024 & Hybrid Classical Quantum Model & Variational Quantum Circuit (VQC) with LSTM \\
            \hline
        \end{tabular}
    }
\end{table}
\mbox{}
\begin{figure}[ht]
    \centering
    \includegraphics[width=0.70\textwidth]{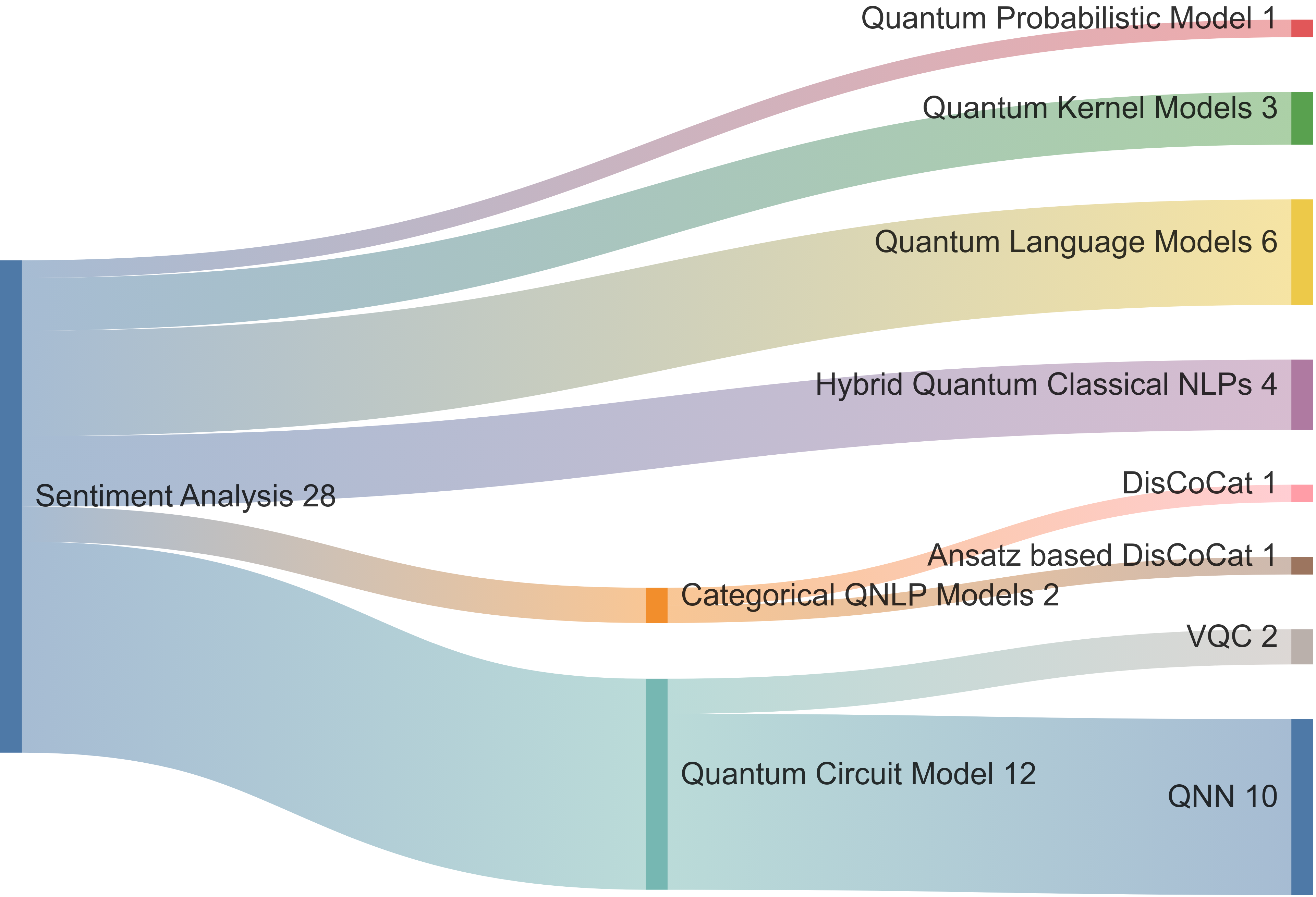}  
    \caption{QNLP Techniques for Sentiment Analysis}
    \label{fig:sentiment_analysis}
\end{figure}
\textbf{Hybrid Classical Quantum Computing Model}
\begin{itemize}
\item Hybrid quantum classical algorithms\cite{10459858} incorporate quantum kernel methods and classifiers based on variational quantum circuits alongside classical dimensionality reduction techniques to enhance performance. The proposed methodology, through the reduction of data dimensionality and the strategic utilisation of the respective strengths of quantum and classical algorithms, consistently surpassed classical methods in sentiment analysis. This was notably demonstrated in the analysis of human emotions and opinions as expressed in large-scale datasets, with empirical validation provided through experiments conducted on datasets in both the English and Bengali languages.
\item A novel methodology, termed Quantum Particle Swarm Optimisation using DL based Sentiment Analysis of Arabic Tweets \cite{al2023quantum}, employs the Quantum Particle Swarm Optimisation algorithm to fine-tune hyper parameters within the Bidirectional Gate Recurrent Unit(BiGRU) classifier. The model attains a notably high accuracy of 98.35\% on Arabic tweets. This study demonstrates that quantum computing, through techniques like QPSO, can optimise deep learning models for sentiment analysis, leading to superior performance compared to traditional approaches.
\item Quantum-based detection of highly semantically similar social bots\cite{liu2023quantum} algorithm identify highly semantic social bots to facilitate the analysis of user-generated content, employ sentiment analysis, and execute data cleaning processes to achieve superior accuracy compared to traditional similarity algorithms.
\item A Quantum Computing based Approach for Sentiment Analysis in Bilateral Conversations \cite{bar2024quantum} represents a hybrid methodology for sentiment analysis within dialogues, integrating both quantum and classical techniques. This approach employs variational quantum circuits complemented by classical data processing and utilises feature extraction through LSTM.
\end{itemize}
In Table ~\ref{tab:SA table}, a summary of the models and techniques employed in the sentiment analysis task is provided.
Figure ~\ref{fig:sentiment_analysis} depicts QNLP techniques for Sentiment Analysis, is a Sankey diagram showing the QNLP methods used in 28 studies. The Quantum Circuit Model is most prevalent, featuring in 12 studies, comprising 10 with Quantum Neural Networks (QNNs) and 2 with Variational Quantum Circuits (VQCs). Other methods include Quantum Language Models (6 studies), Hybrid Quantum-Classical NLPs (4 studies), Quantum Kernel Models (3 studies), Categorical QNLP Models (2 studies) divided into DisCoCat and Ansatz-based DisCoCat methods, and Quantum Probabilistic Models in a single study.


\subsubsection {Text Classification}
Various quantum techniques employed to perform text classification can be depicted below:\\
\textbf{Categorical QNLP Models}
        \begin{itemize}
            \item Distributional Compositional Categorical (DisCoCat) Model:
            A Quantum Natural Language Processing (QNLP) model \cite{metawei2023topic} determines whether two sentences relate to the same topic or not. The proposed model has been evaluated against the traditional tensor network model, revealing an increase in training accuracy by as much as 45\% and validation accuracy by 35\%. These findings indicate that employing strongly entangled ansatz designs leads to the quickest model convergence, considering variations in problem size, the parametrised quantum circuits utilised for model training, and the noise model of the back end quantum simulator.
            \item Another quantum natural language processing (QNLP) model \cite{10.1007/978-3-031-35644-5_7} is used to simulate sentence classification. The study performs a comparison of two optimisation techniques: Simultaneous perturbation stochastic approximation (SPSA) and convergent optimisation via most promising-area stochastic search (COMPASS) for small datasets and simple sentence structures.
            \item Grammar-aware sentence classification on quantum computers \cite{meichanetzidis2023grammar} employs the DisCoCat framework by instantiating sentences as parameterised quantum circuits (PQC) and embedding word meanings as quantum states. The first implementation of an NLP task on an NISQ
processor using the DisCoCat framework. Sentences are instantiated as parameterised quantum circuits; word-meanings
are embedded in quantum states using parameterised quantum-circuits. The circuit's parameters are trained using a classical optimiser for supervised binary classification task. 
\item Running compositional models of meaning on a quantum computer \cite{lorenz2023qnlp} depicts how sentences transformed into quantum circuits incorporating grammar and word meaning are used to perform sentence classification by measuring output states and learning to separate sentence types, all using actual quantum hardware and the Lambeq framework.
        \end{itemize}        
\textbf{Quantum Circuit Models}
\begin{enumerate}
    \item Variational Quantum Algorithms(VQA) 
    \begin{itemize}
    \item
    A hybrid variational quantum classifier\cite{blance2021quantum} intended for current and near-term quantum devices optimises the network parameters by integrating the quantum gradient descent method with the steepest gradient descent.
    \item The variable quantum classifier\cite{katyayan2022supervised} is used to classify the questions in the SelQA dataset consisting of two domains. The investigation mainly concentrated on the implications of the circuit depth in various experiment scenarios. 
    \item A vertical federated learning architecture \cite{yang2022bertmeetsquantumtemporal} employed in variational quantum circuits to augment a pre-trained BERT model for the purpose of text classification. Some layers of the BERT-based decoder are replaced by employing a novel random quantum temporal convolution (QTC) learning framework culminating in the development of the BERT-QTC model, which demonstrates competitive efficacy in intent classification tasks.
       \end{itemize}
\item Quantum Neural Networks(QNNs)
   \begin{itemize}
       \item 
      Quantum Self-Attention Neural Network (QSANN) architecture for text classification tasks \cite{li2024quantum} implements the self-attention mechanism in quantum neural networks and uses a Gaussian-projected quantum self-attention to improve performance on larger datasets. The key element of the Quantum Self-Attention Neural Network (QSANN) is the Quantum Self-Attention Layer (QSAL), where quantum analogous entities of the conventional self-attention constituents, namely queries, keys, and values, are devised. The QSAL derives classical outputs based on quantum principles and incorporates regularisation terms to enhance the network's efficacy.
\item The interpretable complex-valued word embedding technique (ICWE) \cite{shi2021two} designs two end-to-end quantum-inspired deep neural networks (ICWE-QNN and CICWE-QNN) for binary text classification. ICWE uses deep learning methodologies, including the gated recurrent unit (GRU) to extract the sequential information present in sentences, the attention mechanism to emphasise salient words, and convolutional layers to capture local features within the projected matrix.
The second model, CICWE-QNN, further enhances the performance of ICWE-QNN by utilising a 2-D convolutional structure to capture local features of the projected matrix. This model considers non-diagonal values of the matrix, leading to a more comprehensive consideration of real-part and imaginary-part textual features of the quantum mixed system. Compared to the quantum-inspired model CE-Mix, CICWE-QNN demonstrates higher accuracy on four datasets (SST, SUBJ, CR, and MPQA) by avoiding text information loss.
\item A quantum-based KNN algorithm \cite{shi2023naturalnisqmodelquantum} employs fidelity to determine the similarity among two quantum states. This algorithm uses the Control-Swap Test gate as a fidelity estimator.
Quantum computing algorithms are emphasised in their ability to efficiently compute high-dimensional vectors in large tensor product spaces, achieving exponential speed-up over classical counterparts.
\item A quantum classifier\cite{pandey2024quantum} proposes a recurrent quantum neural network to perform classification tasks on text data. 
\item A Quantum-inspired Hierarchical Semantic Interaction Model (QSIM)\cite{gao2025qsim} employs Schmidt decomposition to partition the word semantic space and isolate the fundamental semantic space for the purpose of semantic representation. Sentences are represented as quantum superposition states of words, and the extent of semantic interaction at the word level is quantified through entanglement entropy.

\end{itemize}

\item Quantum Kernel Methods
\begin{itemize}
    \item Quantum Support Vector Machines(QSVM):
    \begin{itemize}
        \item The Quantum Support Vector Machine (QSVM) algorithm \cite{10126672} emphasises the implementation of quantum machine learning techniques to address classification problems. Although the study specifically addresses the coloured-dots dataset, the underlying methodologies can be extrapolated to text classification scenarios. Quantum computing holds promise for advancing text classification in MC and RP data sets through the implementation of quantum algorithms such as the Quantum Support Vector Machine (QSVM). This approach facilitates the attainment of more precise and efficient classification results. Using quantum feature map schemes and oracle functions, quantum computing may enhance the classification accuracy of text data sets by efficiently processing and analysing complex data patterns. Therefore, this has the potential to exceed the performance of conventional machine learning methodologies.
    \item Quantum entanglement \cite{sharma2023role} is employed to enhance the efficacy of quantum kernels in the performance of classification tasks. 
        By classifying the textual data of IMDb movie reviews with a proposed quantum kernel based on linear and fully entangled circuits, the results demonstrate that the fully entangled circuit offers advantages over Classical SVM and linearly entangled circuits when dealing with a large number of features.
\item A lightweight methodology \cite{10547/623794} for optimising word embedding representations aims to reduce the dimensionality of word embeddings derived from larger pre-trained corpora, such as GloVe and FastText, to enhance the performance of quantum-based models, particularly in contexts where complexity and input size represent crucial considerations. The proposed dimensionality reduction demonstrates superior and more efficient performance compared to the state-of-the-art quantum and classical models, as evaluated on the Question Classification and Sentiment Analysis datasets.
   \end{itemize}
    \end{itemize}

\item Hybrid Classical Quantum Computing Model:
\begin{itemize}
    \item 

A fine-tuned quantum pre-trained network \cite{ardeshir2024hybrid}
using the Bidirectional Encoder Representations from Transformers (BERT) model. The mentioned methodology is a classical quantum transfer learning model where pre-trained neural architecture is implemented classically, while the final stages integrate a quantum circuit, quantum measurements, and post-processing facilitating high-precision text classification. 

\item 
A synchronistic framework integrates classical and quantum machine learning methodologies to develop a text classification pipeline. The Quantum Text Classifier (QTC) framework involves pre- and post-processing on classical computers, with the core classification performed using quantum algorithms. This framework has been implemented using IBM's Qiskit library, showcasing the feasibility of quantum text classification\cite{santi2023quantumtextclassifier}
\item A classifier based on quantum computation theory\cite{liu-etal-2013-novel-classifier} views classification as an evolving physical system via basic quantum mechanics. The viability of the quantum classifier is assessed through an experimental analysis performed on two datasets, highlighting the potential of the novel classification approach.
\item  LEXIQL, a novel noise-aware QNLP technique
for text classification on NISQ quantum machines \cite{10.1109/SC41406.2024.00073} addresses the impracticality of the DisCoCat model and introduces the data injection technique to transform textual data into a quantum circuit. Using effective training methods, it can perform text classification.
\end{itemize}
\end{enumerate}

\begin{figure}[ht]
    \centering
    \includegraphics[width=0.75\textwidth]{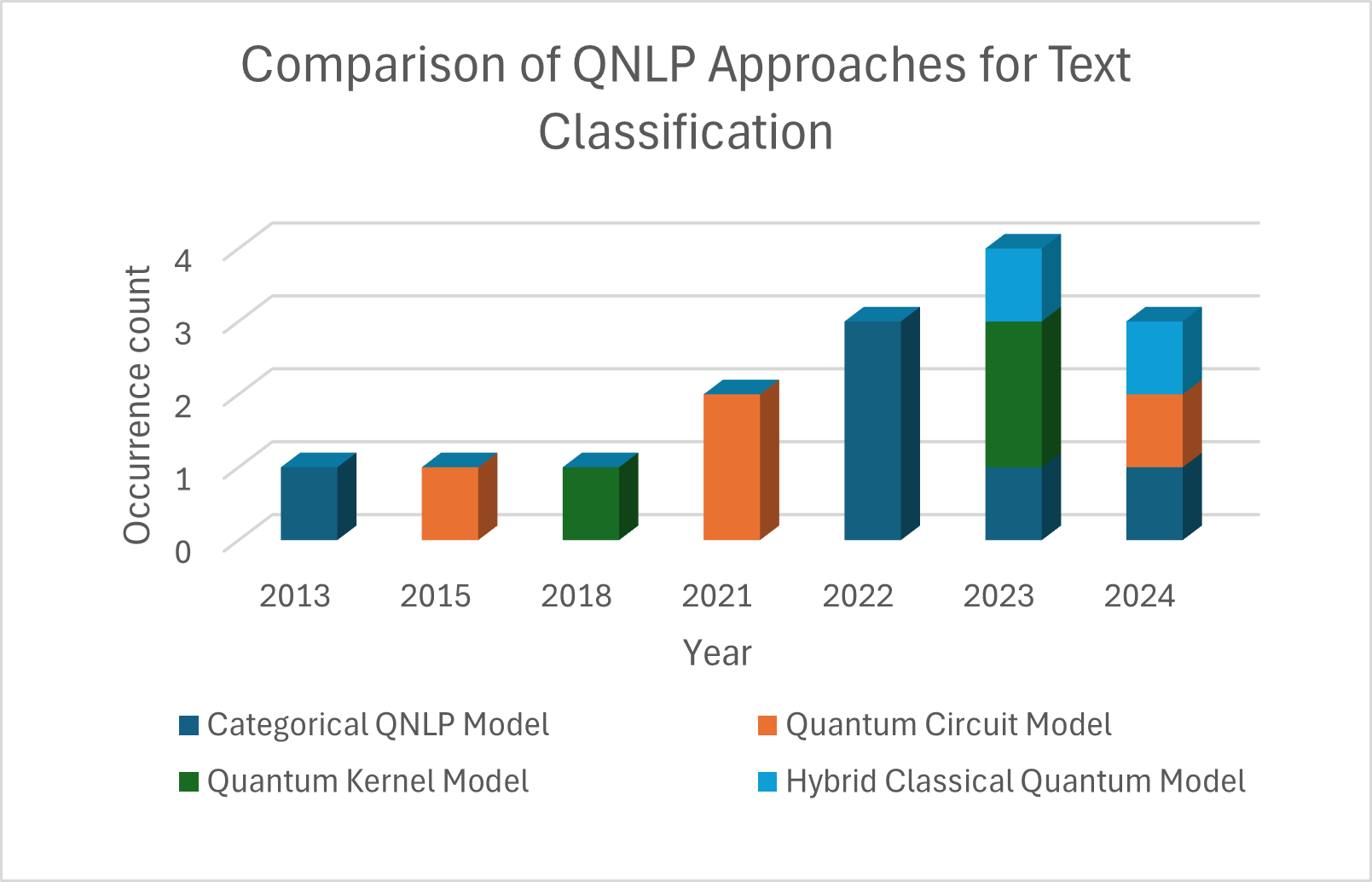}  
    \caption{Text Classification}  \label{fig:text_classification}
\end{figure}

In Table ~\ref{tab:qnlp_textclassif}, a summary of the models and techniques used in the text classification task is provided. Figure ~\ref{fig:text_classification} shows a bar graph of the Quantum Natural Language Processing (QNLP) models used in text classification from 2013 to 2024, with notable growth after 2021. The Quantum Circuit Model (orange) increased notably after 2021, peaking in 2023. Hybrid Classical Quantum Models (blue) grew most in 2022 and 2023, showing rising relevance. The Quantum Kernel Model (green) and Categorical QNLP Model (light blue) appeared less frequently. In 2023, the diversity of the QNLP model was highest, reflecting broader experimentation and integration. This trend highlights the growing exploration and effectiveness of hybrid and circuit-based quantum models in NLP.

\begin{table}[ht]
    \caption{Summary of QNLP Models and Methods in Text Classification}
    \centering
    \resizebox{0.8\textwidth}{!}{  
        \begin{tabular}{|c|c|c|c|}
            \hline
            \textbf{Work} & \textbf{Year} & \textbf{QNLP Model} & \textbf{Methods Used} \\ \hline
            \cite{metawei2023topic} & 2023 & Categorical QNLP Model & DisCoCat \\ \hline
            \cite{10.1007/978-3-031-35644-5_7} & 2022 & Categorical QNLP Model & DisCoCat, SPSA, COMPASS \\ \hline
	\cite{meichanetzidis2023grammar}	&2023	& Categorical QNLP Model	& DisCoCat,  \\ \hline
	\cite{lorenz2023qnlp}	& 2023&	Categorical QNLP Model& 	DisCoCat\\ \hline

            \cite{blance2021quantum} & 2022 & Quantum Circuit Model & Variational Quantum Classifier\\ \hline
            \cite{katyayan2022supervised} & 2022 & Quantum Circuit Model & Variational Quantum Classifier\\ \hline
            \cite{yang2022bertmeetsquantumtemporal} & 2022 & Quantum Circuit Model & Hybrid model using BERT and Quantum Classifier \\ \hline
            \cite{li2024quantum} & 2024 & Quantum Circuit Model & Quantum Neural Network (QNN) \\ \hline
            \cite{shi2021two} & 2021 & Quantum Circuit Model & Quantum Neural Network (QNN) \\ \hline
            \cite{shi2023naturalnisqmodelquantum} & 2015 & Quantum Circuit Model & Quantum k-Nearest Neighbors (QKNN)\\ \hline
\cite{gao2025qsim}	& 2023	& Quantum Circuit Model&	Quantum Neural Network (QNN)\\ \hline
            \cite{pandey2024quantum}	& 2024	& Quantum Circuit Model	& Quantum Recurrent Neural Network (QRNN)\\ \hline
            \cite{10126672} & 2023 & Quantum Kernel methods & Quantum Support Vector Machine (QSVM) \\ \hline
            \cite{sharma2023role} & 2023 & Quantum Kernel methods & QSVM with quantum entanglement features\\ \hline
            \cite{10547/623794} & 2018 & Quantum Kernel methods & Dimensionality reduction in Hilbert space \\ \hline
            \cite{ardeshir2024hybrid} & 2024 & Hybrid Classical Quantum QNLP Model & VQC, Transfer learning \\ \hline
            \cite{santi2023quantumtextclassifier} & 2023 & Hybrid Classical Quantum QNLP Model & Quantum Classifier Framework, Quantum Machine Learning (QML) Algorithms\\ \hline
            \cite{liu-etal-2013-novel-classifier} & 2013 & Hybrid Classical Quantum QNLP Model & Quantum Classifier framework \\ \hline
            \cite{10.1109/SC41406.2024.00073} & 2024 & Hybrid Classical Quantum QNLP Model & Quantum encoder \\ \hline
        \end{tabular}
    }
    \label{tab:qnlp_textclassif}
\end{table}
\smallskip 
\subsubsection{Language Translation}

\begin{enumerate}
    \item {Categorical QNLP Models}
        \begin{itemize}
        \item {Distributional Compositional Categorical (DisCoCat) Model}:
            The study titled 'Toward Quantum Machine Translation of Syntactically Distinct Languages' \cite{abbaszade2023quantummachinetranslationsyntactically} employs Long Short-Term Memory (LSTM) networks for the translation of English and Persian sentences. A variety of optimisation techniques are explored, including stochastic gradient descent (SGD) and combinations with other optimisers; however, the most favourable results are obtained using the Adam optimiser.Shannon entropy is utilised in the optimisation of quantum circuits, yielding promising outcomes.
        \item {Compositional Distributional Circuits (DisCoCirc)} Model: The DisCoCirc framework exemplifies linguistic independence\cite{Waseem_2022} by employing a straightforward translation mechanism between the English and Urdu languages, effectively mitigating grammatical disparities between them. The investigation of the potential of DisCoCirc lies in its emphasis on compositional and generative components rather than specific grammatical conventions.
        \item The space-efficient word embeddings inspired by quantum entanglement \cite{panahi2020word2ketspaceefficientwordembeddings} propose word2ket and word2kets techniques to store word embedding matrix in a highly efficient way. The proposed word embeddings is evaluated for the IWSLT2014 German-to-English machine translation task. 
        \item 
        Application of Quantum Natural Language
Processing for Language Translation \cite{abbaszade2021application} performs the comparison of parameterised quantum circuits(PQC) of two synonymous simple sentences in english and persian language. The proposed protocol is based on quantum long short term memory (Q-LSTM) to perform translation of sentence from English to Persian.

        \item Morphologically enhanced tensorised embedding (MorphTE)\cite{gan2022morphte} represents word embedding as a tensor product of its constituent morpheme vectors, capturing an entangled structure that encodes prior semantic and grammatical knowledge into the embedding learning process.  The experimental results performed on four translation datasets for different
languages show that MorphTE can compress the word embedding parameters about 20 times without compromising performance.
\item Towards Machine Translation with Quantum Computers \cite{vicente2021towards} transforms the English and Spanish
sentences to DisCoCat diagrams which are fed as inputs to 
quantum circuits. The technique preserves the sentence meanings by computing cosine similarity for the two vectorised sentences. 
Table 1 provides an overview for quantum computing techniques used for machine translation.\\
        \end{itemize}
        
\item Quantum Circuit Models

\begin{itemize}
   \item Quantum Neural Networks(QNNs): Machine translation method that combines advantage of QNN \cite{narayan2013machine} and demonstrates better bilingual translation with higher accuracy.  
   \item The QNN based machine translator from Hindi to English, as referenced in \cite{narayan2014quantum}, employs a three-layer QNN architecture for machine translation. The architecture comprises an input, a hidden, and an output layer. A multi-level activation function is employed alongside a methodology grounded in the parts of speech of individual words within the corpus. This approach allows for the interpretation and recognition of corpus patterns in a manner analogous to human cognitive processes.    
\item 
The QNN based machine translator for English to Hindi, as referenced in \cite{narayan2016quantum}, employs a semantically accurate corpus to enhance the translation process by leveraging a QNN. This methodology undertakes the evaluation of 4600 sentences sourced from news media for simulations, achieving noteworthy accuracy metrics. 
\item The implementation of Quantum Natural Language Processing for Language Translation\cite{abbaszade2021application}, using the quantum-based long-short-term memory protocol (QLSTM), facilitates the comparative analysis of synonymous sentences in English and Persian. This approach demonstrates faster convergence with enhanced accuracy compared to classical techniques. Sentence parsing is carried out using two methodologies: the bigraph method and the snake removal method.
    \end{itemize}
\end{enumerate}

\begin{figure}[ht]
    \centering
    \includegraphics[width=0.6\textwidth]{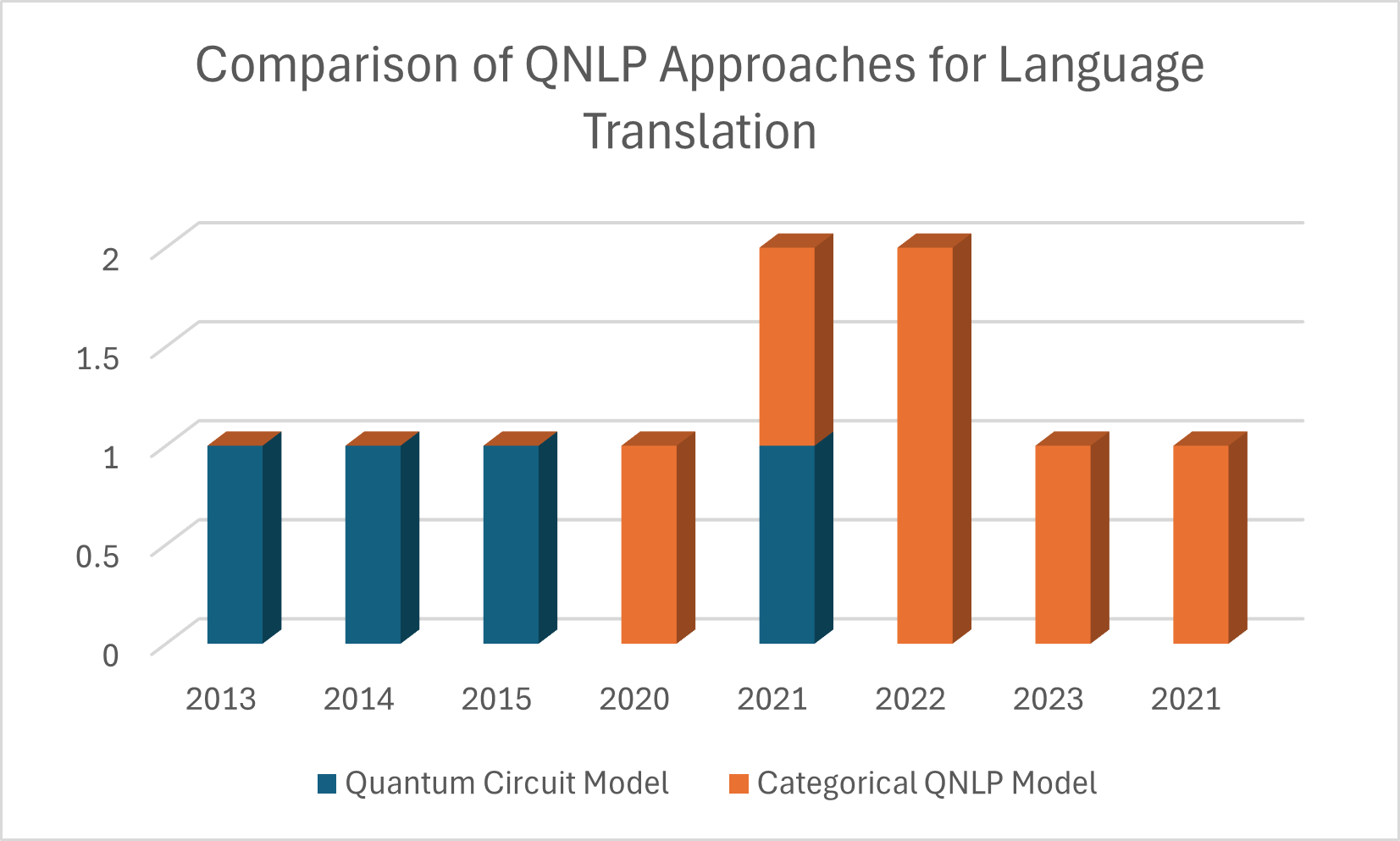}  
    \caption{Language Translation}
    \label{fig:Language_Translation}
\end{figure}

In Table ~\ref{fig:Language_Translation}, a summary of the models, methods used, type of language translation, and data set used in the language translation task is provided.
Figure ~\ref{fig:Language_Translation} compares Quantum Circuit Models and Categorical QNLP Models for language translation between 2013 and 2023. Quantum Circuit Models (blue) maintained steady usage from 2013 to 2021. From 2021, Categorical QNLP Models (orange) gained prominence and surpassed Circuit Models in 2021 and 2022, indicating a shift toward categorical methods for language translation.

Overall, the figure highlights evolving model preferences and growing interest in categorical methods for handling complex NLP tasks like translation in recent years.
\begin{table}[ht]
\centering
\caption{Summary of QNLP Models and Methods used in Language Translation Task}
\label{tab:language translation}
\begin{tabular}{|p{1.05cm}|c|p{2.2cm}|p{2.5cm}|p{2.2cm}|p{2cm}|}
\hline
\textbf{Work} & \textbf{Year} & \textbf{QNLP Model} & \textbf{Methods Used} & \textbf{Language Translation} & \textbf{Dataset} \\
\hline
\cite{narayan2013machine} & 2013 & Quantum Circuit Model & QNN & English to Devnagiri-Hindi & 100 Sentences \\
\hline
\cite{abbaszade2023quantummachinetranslationsyntactically} & 2023 & Categorical QNLP Model & DisCoCat, Shannon entropy & English to Persian & 160 Sentences \\
\hline
\cite{Waseem_2022} & 2022 & Categorical QNLP Model & DisCoCirc & English to Urdu & Not Specified \\
\hline
\cite{narayan2014quantum} & 2014 & Quantum Circuit Model & QNN & Hindi to English & 2600 Sentences \\
\hline
\cite{narayan2016quantum} & 2015 & Quantum Circuit Model & QNN & English to Hindi & 4600 Sentences \\
\hline
\cite{abbaszade2021application} & 2021 & Quantum Circuit Model & QLSTM & English to Persian & Small set of simple sentence pairs \\
\hline
\cite{panahi2020word2ketspaceefficientwordembeddings} & 2020 & Categorical QNLP Model & Tensor Quantum Embedding & German to English & ~160,000 Sentence Pairs \\
\hline
\cite{gan2022morphte} & 2022 & Categorical QNLP Model & Tensor Quantum Embedding & German→English, English→Italian, English→Spanish, English→Russian & De-En: 160K; Others: 1M Pairs Each \\
\hline
\end{tabular}
\end{table}

\subsubsection{Question Answering:}
\begin{figure}[ht]
    \centering
    \includegraphics[width=0.6\textwidth]{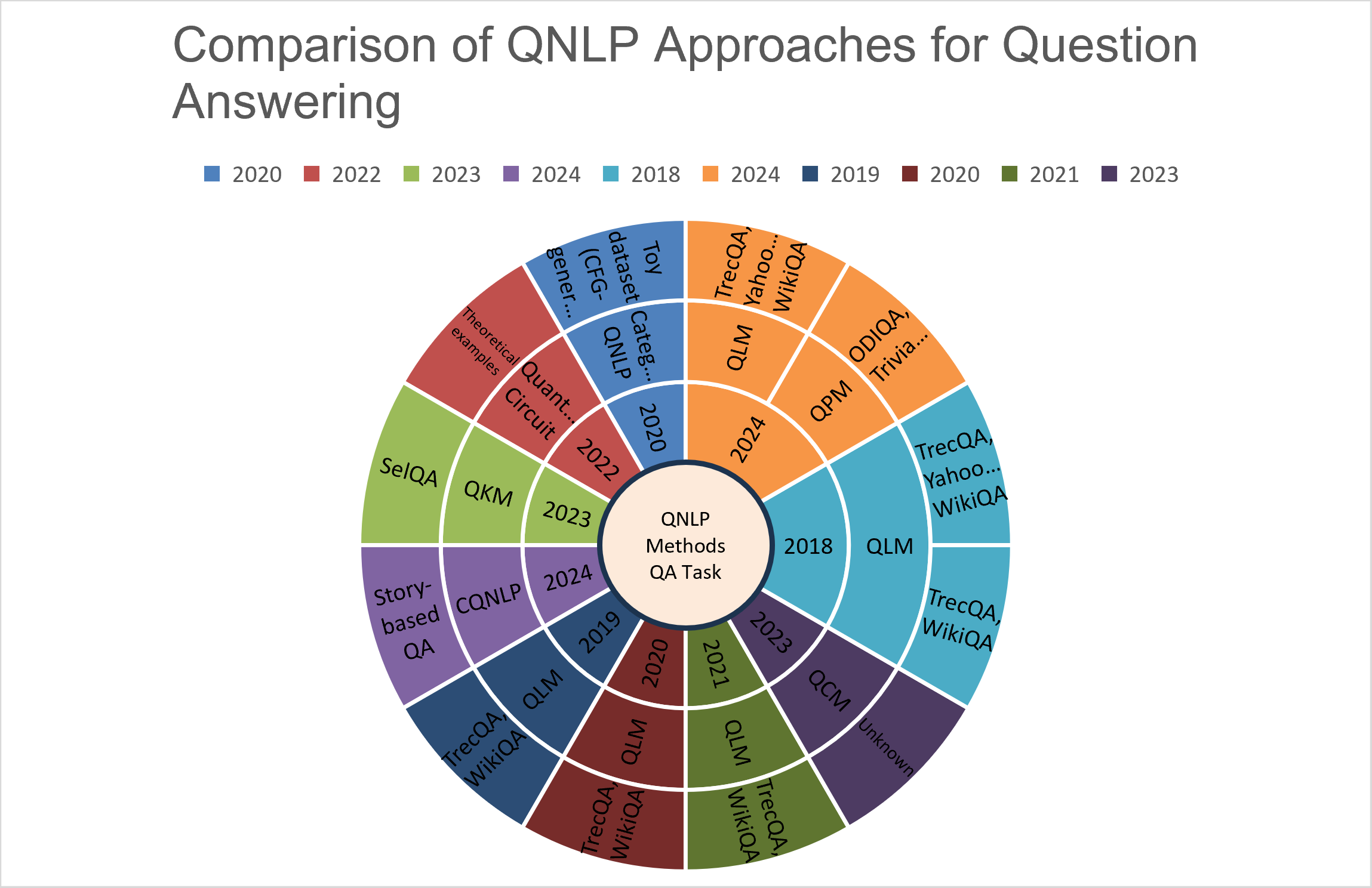}  
    \caption{Question Answering}
    \label{fig:question_answering}
\end{figure}

Quantum computing introduces innovative methodologies for the question answering task (QA), which contribute to the advancement of natural language processing (NLP). Quantum language models(QLMs) and quantum algorithms have demonstrated potential to improve the accuracy of QA systems.
\begin{enumerate}
    \item Categorical QNLP Models
        \begin{itemize}
            \item Distributional Compositional Categorical (DisCoCat) Model: The approach of grammar-aware question answering \cite{Meichanetzidis_2023} on quantum computers involves the instantiation of sentences as parameterised quantum circuits, with word meanings encoded as quantum states. This method integrates grammatical structures, rendering it a feasible option for NISQ devices.
            \item DisCoCirc Model: An experimental execution of the QDisCoCirc model \cite{duneau2024scalableinterpretablequantumnatural} illustrates the potential use of DisCoCirc models in quantum environments to tackle question-answer tasks in specially designed toy datasets. The model's compositional structure facilitates the generalisation to larger instances, while its semantic framework improves interpretability. 

        \end{itemize}
\item Quantum Probabilistic Model
\begin{itemize}
     \item A novel approach \cite{duan2024quantum} integrates the classical mixture method from quantum information theory to improve open-domain question answering. This method employs an effective Quantum Fusion Module (QFM) to proficiently integrate information from individual tokens, thereby aiding the model in achieving a more comprehensive comprehension of questions and passages.
\end{itemize}
\item Quantum Circuit Models

\begin{itemize}
    \item Variational Quantum Algorithms(VQA):
a quantum computing framework using Grover's search algorithm \cite{correia2022quantum} and tensor contraction build word representation as quantum states and handles syntactic ambiguities. 
    \item By integrating semantic classification and tensor operations\cite{han2023quantum}, quantum computing resources can be optimised. This method given current resource constraints explores the potential for the question-answer task. 
   \end{itemize}
\item Quantum Kernel Model
\begin{itemize}
    \item The question classification is performed using the SelQA dataset \cite{Katyayan_2023} using quantum-based classifier algorithms, namely the quantum support vector machine (QSVM) and the variational quantum classifier (VQC). Question classification is performed using both classifiers and in similar environments to study the impacts and compare the results of both classifiers.
\end{itemize}
\item Quantum Language Models
\begin{itemize}
\item
In \cite{li-etal-2019-cnm} a quantum-inspired neural model is adopted that extends quantum language models with complex-valued embeddings and quantum principles. The effectiveness of the model for the question-answer task is demonstrated. 
\item
Quantum Language Model with Entanglement Embedding \cite{chen2021quantum} converts word sequences into entangled pure states captures non-classical correlations to enhance performance on QA datasets.
\item The Neural Network-based Quantum-like Language Model (NNQLM)\cite{Zhang_Niu_Su_Wang_Ma_Song_2018}, designed for question-response, uses word embeddings along with a density matrix. The density matrix represents questions or answers that encapsulate a mix of semantic subspaces. This approach is incorporated into neural network frameworks for efficient joint training.
\item
The approach to language modelling, which draws inspiration from the quantum many-body wave function (QMWF) \cite{10.1145/3269206.3271723} , represents intricate interactions among words, each associated with multiple semantic basis vectors to account for various meanings and concepts, through the adoption of the tensor product. Additionally, this approach underscores the significance of convolutional neural networks (CNNs) in the QMWF language modelling framework.
\item The study titled "A Quantum Expectation value-based language model with application to question answering"\cite{zhao2020quantum} conceptualises words and sentences as distinct observables within a quantum framework, employing a unified density matrix to evaluate joint question-answer observables. Within this framework, the compatibility score of a question-answer pair is inherently interpreted as the quantum expectation value of the respective joint question-answer observable.
\item 
The Quantum-inspired Language Model Based on Unitary Transformation (QLM-UT)\cite{fan2024quantum} utilises a unitary transformation module to comprehensively encapsulate the dynamic evolution of sentence semantics. This real-valued, quantum-inspired model is particularly applied to tasks in Question-Answering (QA) and text classification. QLM-UT is claimed to necessitate fewer parameters for sentence representation when compared to its complex-valued quantum-inspired counterparts.
\end{itemize}

\end{enumerate}
Table ~\ref{tab:qnlp_qa_summary} provides an overview of the QNLP models, detailing the methodologies applied and the data sets used for the task of answering questions. Figure ~\ref{fig:question_answering} shows a sunburst chart of QNLP methods for the QA task across different datasets and years. The centre is QA, branching to methods like QLM, CQNLP, QKRM, VQC, etc. The next layer shows the datasets (e.g., TrecQA, WikiQA, YahooQA, Story-based QA) used, and the outer ring shows the years from 2018 to 2024. The figure illustrates the evolution and diversity of QNLP in QA, highlighting increased adoption and experimentation with quantum models across various datasets.

\begin{table}[ht]
\centering
\caption{Summary of QNLP Models and Methods used in Question Answering Task}
\label{tab:qnlp_qa_summary}
\begin{tabular}{|p{1.0cm}|c|p{3.2cm}|p{6cm}|p{4.5cm}|}
\hline
\textbf{Work} & \textbf{Year} & \textbf{QNLP Model} & \textbf{Quantum Implementation / Method} & \textbf{Datasets} \\
\hline
\cite{Zhang_Niu_Su_Wang_Ma_Song_2018} & 2018 & Quantum Language Model & Quantum-inspired superposition and measurement modeling & TRECQA, WIKIQA \\
\hline
\cite{zhang2018quantum} & 2018 & Quantum Language Model & Tensor Product Encoding, Quantum-inspired CNN & TRECQA, YAHOOQA, WIKIQA \\
\hline
\cite{li-etal-2019-cnm} & 2019 & Quantum Language Model & Complex Hilbert Space, Gleason Measurement & TRECQA, WIKIQA \\
\hline
\cite{Meichanetzidis_2023} & 2020 & Categorical QNLP Model & DisCoCat Model, Parameterised Quantum Circuits & Toy dataset (CFG-generated) \\
\hline
\cite{zhao2020quantum} & 2020 & Quantum Language Model & Global Density Matrix, Quantum Expectation Value & TRECQA, WIKIQA \\
\hline
\cite{chen2021quantum} & 2021 & Quantum Language Model & Entanglement Embedding + CNN & TRECQA, WIKIQA \\
\hline
\cite{correia2022quantum} & 2022 & Quantum Circuit Model & Tensor Contraction, Grover's Algorithm & Theoretical examples \\
\hline
\cite{Katyayan_2023} & 2023 & Quantum Kernel Method & QSVM, VQC (Quantum SVM and Variational Quantum Classifier) & SelQA \\
\hline
\cite{han2023quantum} & 2023 & Quantum Circuit Model & Quantum Semantic Coding & Unknown \\
\hline
\cite{duneau2024scalableinterpretablequantumnatural} & 2024 & Categorical QNLP Model & QDisCoCirc, Trapped Ion Hardware & Story-based QA (toy dataset) \\
\hline
\cite{duan2024quantum} & 2024 & Quantum Probabilistic Model & Quantum Probabilistic Fusion Module & ODIQA, TRIVIAQA \\
\hline
\cite{fan2024quantum} & 2024 & Quantum Language Model & Unitary Transformation Mechanism & TRECQA, YAHOOQA, WIKIQA \\
\hline
\end{tabular}
\end{table}


\subsubsection{Text Summarisation}
The quantum-inspired algorithm provides promising methodologies for text summarisation, effectively addressing the challenges inherent in the summarisation of multiple documents. By integrating quantum computing principles into text summarisation, the sentence extraction process is optimised, thereby enhancing computational efficiency. 

\begin{figure}[ht]
    \centering
\includegraphics[width=0.6\textwidth]{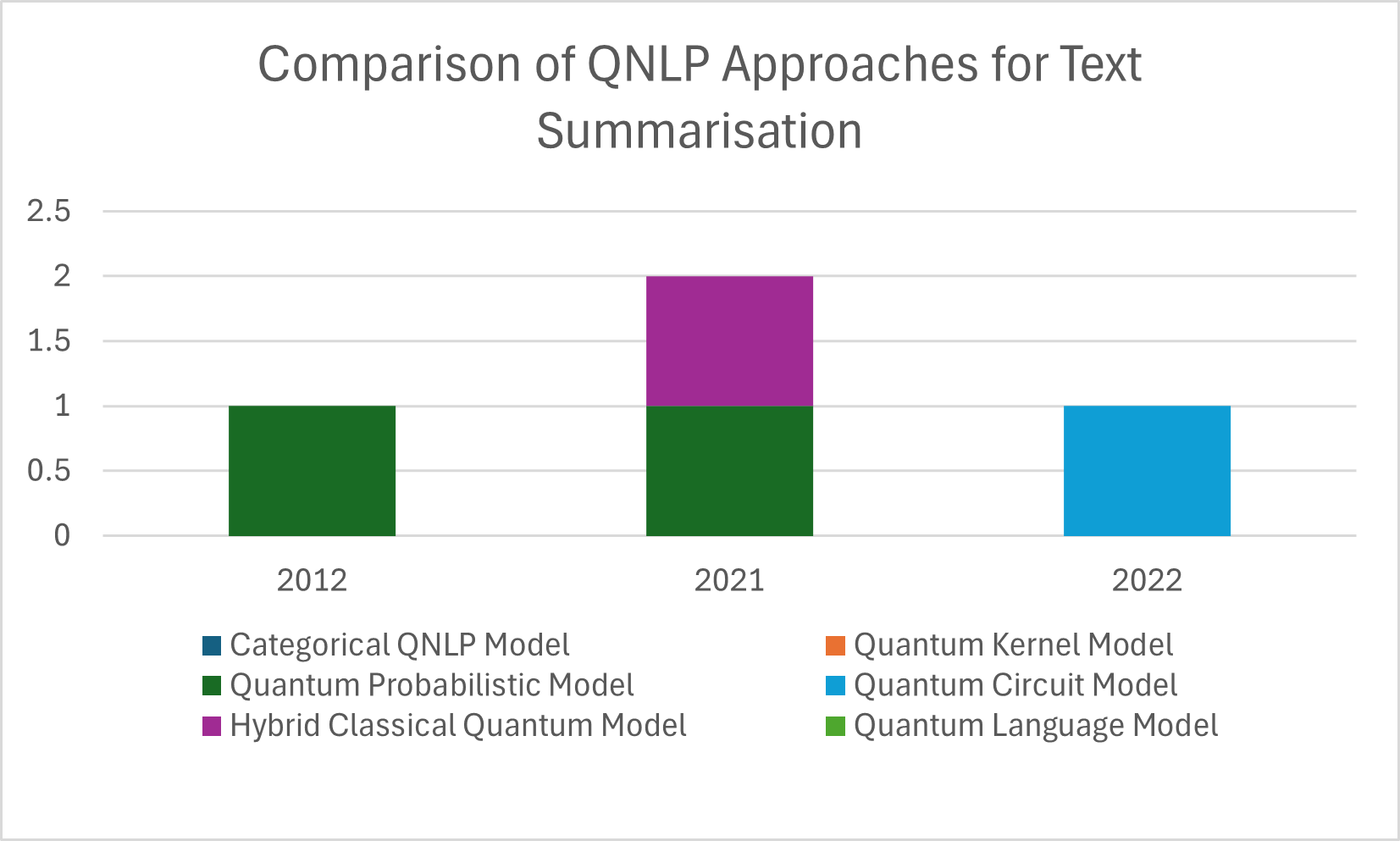}  
    \caption{Text Summarisation}
    \label{fig:text_summarisation}
\end{figure}

\begin{enumerate}
        
\item Quantum Probabilistic Model:
\begin{itemize}
    \item 
The Quantum Information Access (QIA) framework\cite{piwowarski2012using} enhances latent semantic analysis methodologies by incorporating principles of the quantum physics formalism, thus improving the discernment of themes and the extraction of critical sentences in tasks involving the summarisation of multiple documents.
\item A quantum inspired genetic algorithm (QGA) \cite{chettah2021quantum} performs the extraction of single document summarisation. The proposed genetic algorithm is used as an optimiser to search for the best combination of sentences and its performance is evaluated compared to the latest methods.
\end{itemize}
\item Quantum Circuit Models
An additional methodology, the Quantum Alternating Operator Ansatz algorithm \cite{niroula2022constrained}, in conjunction with a Hamming weight-preserving algorithm, facilitates the resolution of the extractive summarisation constrained-optimisation problem.
\item Hybrid Classical Quantum Computing Model:
The MTSQIGA \cite{mojrian2021novel} proposes a system for multi-document summarisation of text using a quantum-based genetic algorithm that incorporates quantum-based measurement techniques and a quantum rotation gate to obtain an optimal solution. 
Table ~\ref{tab:methods_datasets} summarises the QNLP model, methods used, and dataset for the Text summarisation task. Figure ~\ref{fig:text_summarisation} illustrates the QNLP approaches for text summarisation in 2012, 2021, and 2022. In 2012, Quantum Language Models dominated. By 2021, both Quantum Language and Hybrid Classical Quantum Models were used. In 2022, Categorical QNLP models emerged, indicating increased experimentation and diversification over time.
\end{enumerate}
\begin{table}[ht]
\label{tab:text summarisation}
  \caption{Summary of QNLP Models and Methods Used in Text Summarisation}
    \centering
    \resizebox{0.8\textwidth}{!}{  
        \begin{tabular}{|c|c|c|c|c|}
            \hline
            \textbf{Work} & \textbf{Year} & \textbf{QNLP Model} & \textbf{Methods used} & \textbf{Dataset} \\ \hline
            \cite{piwowarski2012using}     & 2012   & Quantum Probabilistic Model (QPM) & Quantum Information Access (QIA) framework  & DUC 2005, DUC 2006, DUC 2007 \\ \hline
            \cite{chettah2021quantum}  & 2021   & Quantum Probabilistic Model (QPM) & Quantum genetic algorithm & DUC 2001 \& DUC 2002 \\ \hline
            \cite{mojrian2021novel}         & 2021   & Hybrid Classical-Quantum Model & Multi-document Text Summarisation (MTSQIGA)  & DUC 2005, DUC 2007 \\ \hline
            \cite{niroula2022constrained} & 2022   & Quantum Circuit Model (QCM) & Quantum Alternating Operator Ansatz algorithm & CNN/DailyMail dataset (300,000+ articles) \\ \hline            
        \end{tabular}
    }
      \label{tab:methods_datasets}
\end{table}

\subsection{In what ways might quantum optimisation algorithms, be employed to enhance hyper parameter tuning or other optimisation tasks within natural language processing models?}
Quantum optimisation algorithms offer innovative solutions for hyperparameter tuning in NLP models. These algorithms leverage quantum computing capabilities to enhance the efficiency and efficacy of optimisation tasks, particularly in dynamic environments.\\
\textbf{Quantum-Accelerated Hyper parameter Tuning}
\begin{itemize}
    \item Dynamic Optimisation: Quantum Accelerated Hyperparameter Tuning (QAHT) \cite{ravikumar2024quantum} Harnesses Quantum Neural Networks (QNN) to dynamically optimise hyper parameters in real time, effectively addressing the challenges posed by changing data distributions in NLP tasks. QAHT improves model performance by ensuring adaptability to shifts in data sentiment, thereby expediting the tuning process.
    \item Another quantum-based alternating operator ansatz algorithm \cite{niroula2022constrained} helps to solve the extractive summarisation constrained optimisation problem.

\end{itemize}
\textbf{Quantum Annealing:} 
\begin{itemize}
\item 
A Quantum Annealing Instance Selection Approach for the Fine-Tuning of Transformers\cite{pasin2024quantum} illustrates the capability to diminish the size of training data sets, whilst preserving the performance of transformer models and expediting the fine-tuning process. The technique yields results comparable to those of state-of-the-art instance selection systems, thereby demonstrating the QC potential. Despite the shortcomings of quantum annealer that currently affect the practical outcomes, it is anticipated that advancements in quantum hardware will enhance the system's efficacy.
\end{itemize}
\textbf{Hybrid Quantum-Classical Approaches}
\begin{itemize}
    \item The model for predicting hyperparameter performance using a hybrid quantum-classical approach, as discussed in \cite{wulff2024distributed}, combines conventional GPU training with quantum-enhanced support vector regression. This integration estimates hyperparameter performance and achieves approximately 9\% resource savings while maintaining or enhancing accuracy. This technique facilitates automation by integrating communication channels between classical and quantum systems, thus streamlining hyper parameter optimisation process. 
\end{itemize}

\section{SUMMARY OF FINDINGS}
After reviewing the selected literature and answering the survey questions, several key points emerged.

\subsection{ Quantum computing for natural language processing}

Quantum Natural Language Processing (QNLP) and Traditional NLP While Natural Language Processing (NLP) constitutes a long-established field within artificial intelligence, dedicated to equipping machines with the capacity to interpret human language through specific computational strategies, Quantum Natural Language Processing (QNLP) employs quantum computing principles to realise feasible solutions that exhibit quantum advantage. NLP relies primarily on classical algorithms and machine learning techniques, such as deep learning, to analyse linguistic data, which has led to substantial advancements in NLP tasks. However, NLP systems may encounter difficulties with complex queries and extensive datasets due to computational constraints. Quantum Natural Language Processing (QNLP) seeks to overcome these challenges through the use of distinct quantum attributes, specifically superposition and entanglement, which are capable of significantly enhancing the speed of processing linguistic tasks and improving the understanding of intricate queries. Thus, it could culminate in more efficient algorithms that can handle human language intricacies better than conventional NLP systems. Therefore, the key difference lies in QNLP's capability to revolutionise language processing by harnessing quantum technologies, distinguishing it from conventional NLP techniques. 

Quantum superposition and entanglement represent fundamental principles integral to the advancement of QC. Superposition enables parallel computations by allowing quantum bits to coexist in multiple states concurrently, while entanglement offers a distinct form of correlation between qubits that can be tuned to enhance the computational power. These principles substantially enhance the capability of quantum computers to explore a vast array of prospective solutions, thereby facilitating the effective resolution of complex problems in a manner contrasting with classical computing methods. Although quantum superposition and entanglement offer notable advantages within the realm of quantum computing, they also present intricacies that require proficient management to fully realise the potential inherent in quantum computing. With the advancement of quantum hardware and the refinement of quantum error correction techniques, the effective utilisation of quantum phenomena can be achieved.

The quantum advantage is in achieving quadratic speed in tasks performed on quantum computers than on classical computers. The first NLP experiment conducted on quantum hardware indicates substantial evidence that the realisation of QNLP is achievable\cite{lorenz2023qnlp}. Grover's algorithm \cite{kwiat2000grover} explored for NLP tasks such as question answering \cite{correia2022groversalgorithmquestionanswering} demonstrates results with quadratic speedup. 
\subsection {Integrating Quantum Encoding with NLP }
Quantum encoding constitutes a crucial component in the advancement of quantum computing and data storage technologies, providing trans formative capabilities for a wide array of applications. Basically, it performs transforms classical data to quantum states, thus fostering execution of complex quantum algorithms and improving data processing. 

The Quantum Text Teleportation Protocol (QTTP) \cite{karthik2022quantum}uses quantum teleportation principles to encode text into quantum states and integrates with the Huffman coding for secure transmission, thus allowing efficient data handling in quantum communication.  
Quantum entanglement encoding techniques\cite{Chen_2023},\cite{yu2020quantum} by considering the correlations among the adjacent words in the input sequence generate quantum states. Although the degree of entanglement of the word states is not quantified, this is likely to impact the interpret ability of the models.

Quantum probability-based encoding\cite{yan2021quantum} effectively captures intricate relationships and the semantics of data to produce quantum states, which holds significance in the domain of document classification tasks. This research suggests the necessity for additional investigation into the incorporation of quantum principles with graph neural networks to enhance document representation and classification. 

The amplitude encoding technique \cite{ghosh2021encodingclassicaldataquantum} incorporates the amplitude encoded feature map to generate quantum states without requiring Qubit for each word.  The study determines the issue of scalability and the need to manage more extensive data sets. 

The encoding of textual information through quantum circuits employs variational quantum circuits\cite{melnikov2023quantum} to produce quantum states. Additional techniques are incorporated to perform optimisation and avoid the barren plateau problem. The research gap identifies the need for future empirical verification of the unconditional quadratic speedup over any classical algorithm in certain situations. 
Quantum measurement-based encoding schemes\cite{huertasrosero2008characterisingerasingtheoreticalframework}\cite{correia2022quantum} incorporate lexical measurements and tensor contractions in quantum states by addressing syntactically ambiguous expressions. These techniques lack a detailed exploration of lexical measurements in the performance of information retrieval systems. 

\subsection {QNLP Models applied to NLP tasks}

The existing literature has been extensively reviewed to conclude on the general categorisation of QNLP models. Based on the utilisation of quantum computing principles and their computational approach, six general categories of QNLP models are categorised as: 
\begin{enumerate}
    \item Categorical QNLP Model
    \item Quantum Probabilistic Model 
    \item Quantum Circuit Model
    \item Quantum Kernel Model 
    \item Quantum Language Model 
    \item Hybrid Classical Quantum Computing Model 
    \end{enumerate}

\begin{figure}[ht]
    \centering
    \begin{minipage}{0.7\textwidth}  
        \centering
        \includegraphics[width=\textwidth]{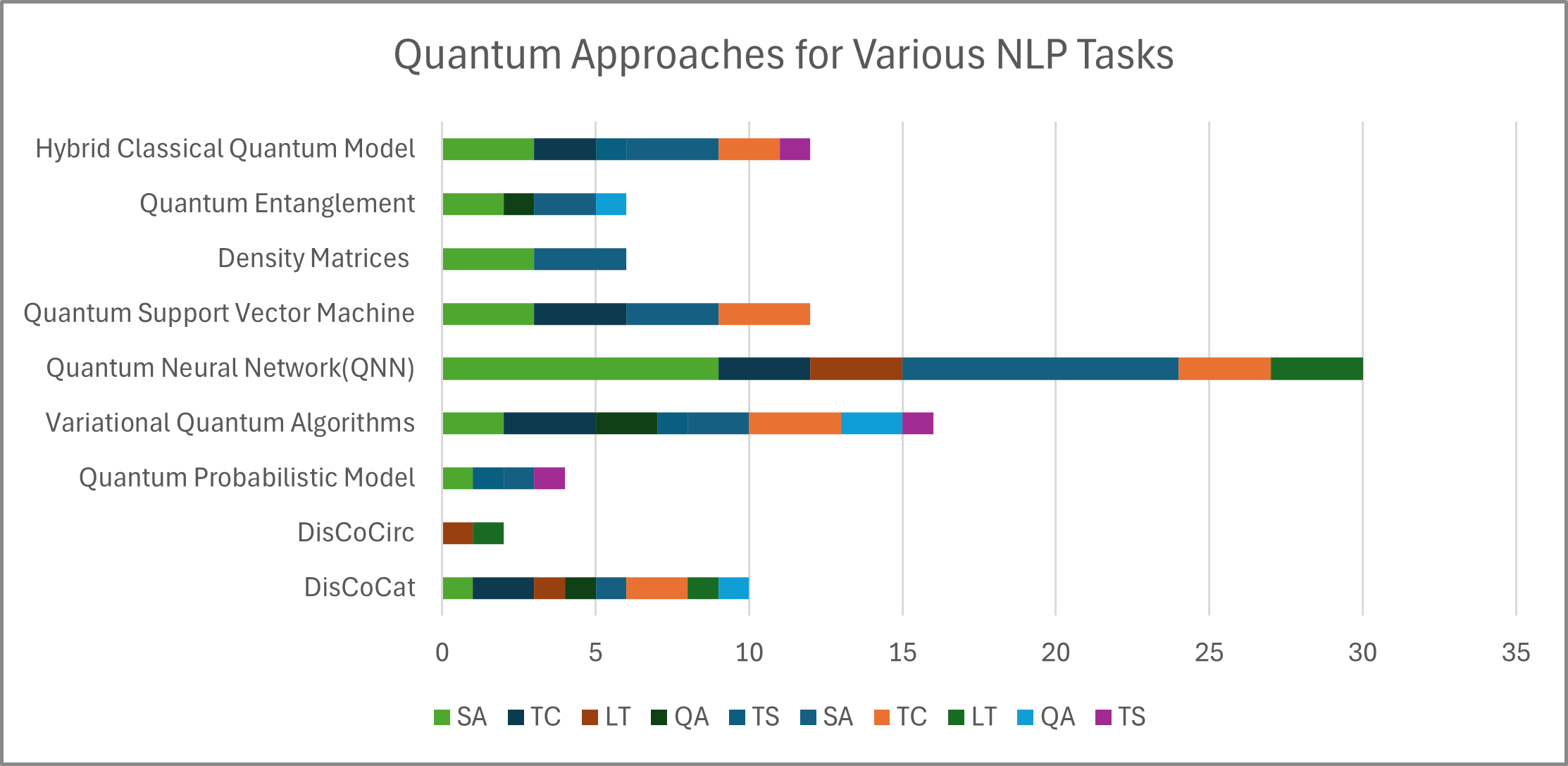}  
        \caption{NLP Task}
        \label{fig:NLPClustered}
    \end{minipage}

   \mbox{}

    \begin{minipage}{0.7\textwidth}
        \centering
        \includegraphics[width=\textwidth]{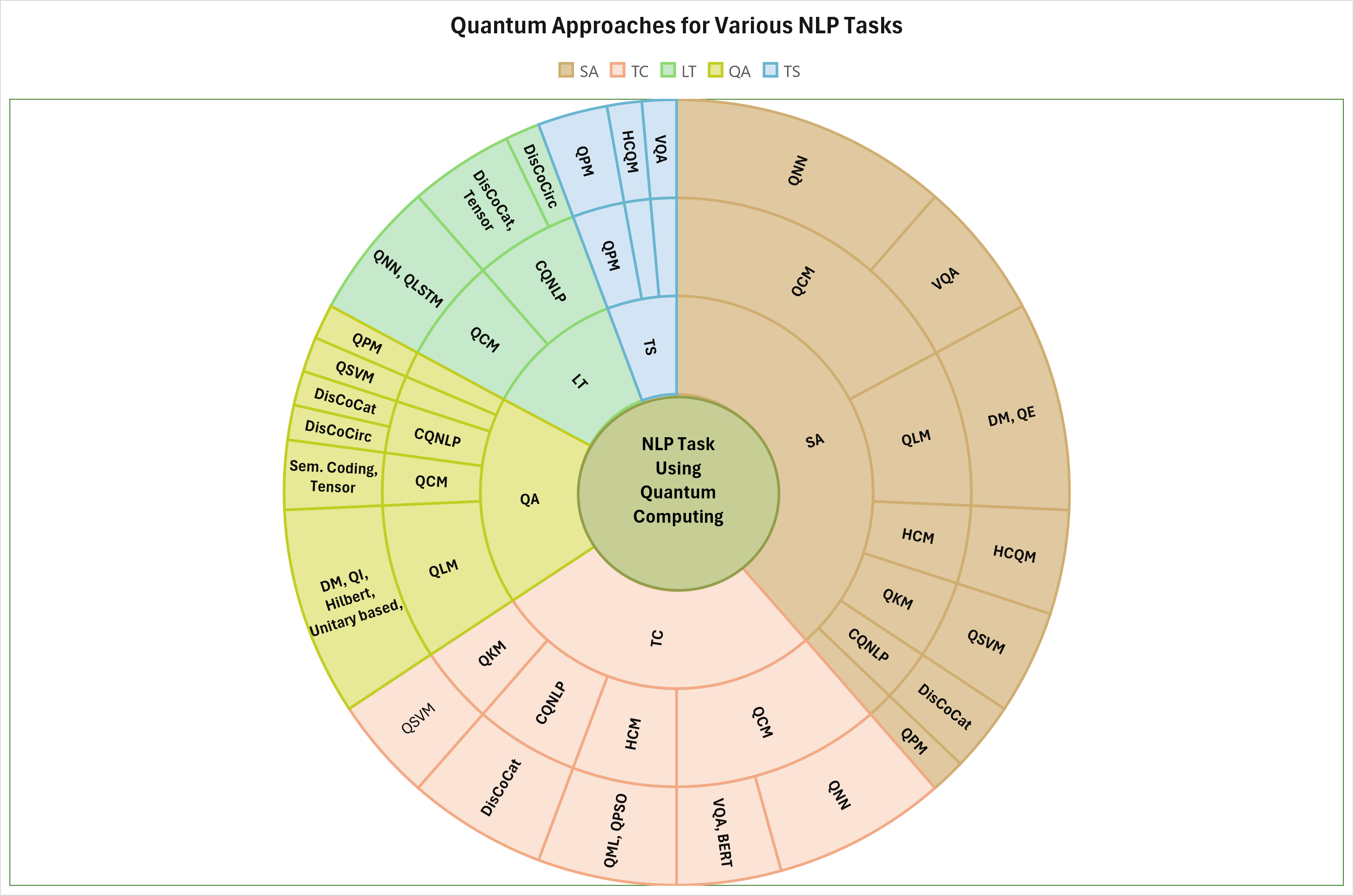}  
        \caption{NLP Tasks}
        \label{fig:NLP Task}
    \end{minipage}
\end{figure}

Prevalent NLP tasks implemented using quantum computing are reviewed in SQ 4 and, depending on their implementations, these tasks are categorised into the above-identified QNLP models. Table 8 summarises the QNLP models and the specific methods adopted for the NLP tasks. It is observed that SA and TC are the tasks that are commonly researched.Quantum circuit models, encompassing variational quantum circuits and quantum neural networks, represent the most extensively utilised QNLP framework across a variety of NLP tasks. It is emphasised that QNLP approaches are still limited, mainly used on small datasets, with only a few models (like QCM and HCM) explored extensively, indicating the need for further research to scale quantum NLP methods. Fig.9. depicts the clustered bar chart representing various QNLP models along with their respective application within a spectrum of NLP tasks, which are highlighted by different colours for each NLP task. Sentiment analysis is represented in blue, text classification is in orange, language translation is in green, question answering is in light blue, and text summarisation is in purple. The following observations were made: 
\begin{itemize} 
\item Prevalence of QNN: Amidst a diverse array of natural language processing tasks, the QNN model exhibits extensive applicability, particularly within the domains of sentiment analysis and language translation. 
\item Versatile Use of Variational Quantum Algorithms : A well-distributed presence across various NLP tasks, encompassing Sentiment Analysis, Question Answering, and Text Summarisation, is exhibited by Variational quantum algorithms. 
\item Minimal use of DisCoCirc and Quantum Probabilistic Models – A limited adoption of DisCoCirc and Quantum Probabilistic models signifies that these frameworks are not thoroughly examined or not extensively utilised in practical NLP applications.
\item The Hybrid Classical-Quantum Models: These models demonstrate moderate engagement across Sentiment Analysis (SA), Text Summarisation (TS), and Text Classification (TC), implying that hybrid methodologies could be a viable conduit toward fully realised quantum NLP models.
\item Quantum Support Vector Machines: QSVM shows a significant prevalence for NLP tasks such as SA and TC. 
\end{itemize}
 Figure ~\ref{fig:NLPClustered} presents a bar graph depicting the distribution of various quantum approaches applied to various natural language processing (NLP) tasks, including Sentiment Analysis (SA), Text Classification (TC), Language Translation (LT), Question Answering (QA) and Text Summarisation (TS). Among the featured methodologies, Quantum Neural Networks (QNN) and Variational Quantum Algorithms are identified as the most widely employed across a wide range of tasks, with QNNs demonstrating the highest overall frequency of implementation.
 Figure ~\ref{fig:NLP Task} illustrates the landscape of quantum computing techniques used in various natural language processing tasks, using a circular chart that categorises tasks such as sentiment analysis, text classification, language translation, question answering and text summarisation. It further highlights methodologies including CQNLP, QCM, QLM, QPM, QKM, and HQM.
Table ~\ref{fig:NLP Task} illustrates the application of QNLP models across various NLP tasks.
\begin{table}[ht]
\label{tab:NLP task}
\centering
\small 
\begin{tabularx}{\textwidth}{|l|l|X|c|l|}
\hline
\textbf{NLP Task} & \textbf{QNLP Model} & \textbf{Methods Used} & \textbf{Approach Count} & \textbf{References} \\
\hline
Sentiment Analysis & CQNLP & DisCoCat & 1 & \cite{9951286} \\
\hline
Sentiment Analysis & QPM & Social bot, quantum similarity alg. & 1 & \cite{liu2023quantum} \\
\hline
Sentiment Analysis & QCM & VQA & 2 & \cite{Joshi_2021}, \cite{singh2022emotion} \\
\hline
Sentiment Analysis & QCM & DQLSTM, CQLSTM, QFNN, Complex valued QNN, Quantum enhanced transformer, Multi modal sentiment analysis, quantum measurements, Complex valued QNN, DMATT, Multi modal sentiment analysis & 9 & \cite{hou2022realization}, \cite{chu2024effective}, \cite{tiwari2024quantum}, \cite{lai2023quantum}, \cite{di2022dawn}, \cite{zhang2020quantum}, \cite{li2021quantum}, \cite{8995180}, \cite{gkoumas2021quantumcognitivelymotivateddecision} \\
\hline
Sentiment Analysis & QKM & QSVM & 3 & \cite{9996813}, \cite{omar2023quantum}, \cite{sharma2023role} \\
\hline
Sentiment Analysis & QLM & DM & 3 & \cite{wang2023quantum}, \cite{zhang2019quantum}, \cite{zhang2018quantum} \\
\hline
Sentiment Analysis & QLM & Aspect-Based Sentiment Analysis (ABSA), Quantum Projection \& entanglement techniques & 2 & \cite{zhao2022quantum}, \cite{Zhao_Wan_Qi_2024}, \cite{wang2023quantum} \\
\hline
Sentiment Analysis & HCM & Quantum Kernel methods and PCA, Quantum Particle Swarm Optimisation, Semantically similar social bots & 3 & \cite{10459858}, \cite{al2023quantum}, \cite{liu2023quantum} \\
\hline
Text classification & CQNLP & DisCoCat & 2 & \cite{metawei2023topic}, \cite{10.1007/978-3-031-35644-5_7} \\
\hline
Text classification & QCM & VQA & 3 & \cite{blance2021quantum}, \cite{katyayan2022supervised}, \cite{yang2022bertmeetsquantumtemporal} \\
\hline
Text classification & QCM & BERT, QNN, QKNN & 3 & \cite{li2024quantum}, \cite{shi2021two}, \cite{shi2023naturalnisqmodelquantum} \\
\hline
Text classification & QKM & QSVM, Dimensionality reduction & 3 & \cite{10126672}, \cite{sharma2023role}, \cite{10547/623794} \\
\hline
Text classification & HCM & VQC and Transfer learning, Quantum Classifier framework, QML Alg, data injection & 2 & \cite{ardeshir2024hybrid}, \cite{santi2023quantumtextclassifier}, \cite{liu-etal-2013-novel-classifier}, \cite{10.1109/SC41406.2024.00073} \\
\hline
Language translation & CQNLP & DisCoCat & 1 & \cite{abbaszade2023quantummachinetranslationsyntactically} \\
\hline
Language translation & CQNLP & DisCoCirc & 1 & \cite{Waseem_2022} \\
\hline
Language translation & QCM & QNN, QLSTM & 3 & \cite{narayan2014quantum}, \cite{narayan2016quantum}, \cite{abbaszade2021application} \\
\hline
Question answering & CQNLP & DisCoCat & 1 & \cite{Meichanetzidis_2023} \\
\hline
Question answering & QCM & Tensor Contraction, Quantum Semantic Coding & 2 & \cite{correia2022quantum}, \cite{han2023quantum} \\
\hline
Question answering & QLM & Quantum entanglement embedding & 1 & \cite{chen2021quantum} \\
\hline
Text summarisation & QPM & Quantum Information Access (QIA) framework & 1 & \cite{piwowarski2012using} \\
\hline
Text summarisation & QCM & Quantum Alternating Operator Ansatz algorithm & 1 & \cite{niroula2022constrained} \\
\hline
Text summarisation & HCM & Multi-document Text Summarisation system using Quantum-Inspired Genetic Algorithm: MTSQIGA & 1 & \cite{mojrian2021novel} \\
\hline
\end{tabularx}
\caption{Summary of NLP Tasks, QNLP Models, Methods Used}
\label{tab:nlp_tasks}
\end{table}
\subsection {Quantum optimisation techniques and QNLP models }
A suite of dynamic optimisation methodologies, specifically Quantum Accelerated Hyperparameter Tuning\cite{ravikumar2024quantum} and the Quantum Alternating Operator Ansatz algorithm utilising a Hamming weight-preserving XY mixer (XY-QAOA)\cite{niroula2022constrained}, iteratively optimises parameters to achieve optimal solutions for dynamic NLP models. Moreover, the quantum annealing technique\cite{pasin2024quantum} is shown to effectively diminish the size of the training data set and enhance the process of model fine-tuning.The distributed hybrid quantum-classical framework for quantum optimisation performs hyperparameter tuning by integrating classical GPU training with quantum trained support vector regression (QT-SVR). \\
These optimisation techniques enable hyper parameter tuning process by harnessing quantum computing capabilities, offering applications in areas such as summarisation and information retrieval systems.  

\section{CONCLUSION}
Using quantum computing principles, QNLP enhances the processing capabilities of NLP tasks. The initial experiment involving natural language processing on quantum hardware offers substantial evidence supporting the potential realisation of quantum NLP \cite{lorenz2023qnlp}. From the available literature, we have proposed a classification of QNLP models. Based on our analysis of survey question SQ 4, it can be concluded that there is a growing interest in applying quantum computing to NLP tasks with sentiment analysis and text classification as mostly researched areas.
 Quantum Neural Networks are the most thoroughly examined models in QNLP, with extensive applicability across various NLP tasks. Nevertheless, alternative frameworks such as variational quantum models and hybrid classical quantum models exhibit potential in addressing an array of NLP challenges. The minimal use of DisCoCirc and Quantum Probabilistic models highlights a research gap that warrants further exploration.

Quantum encoding techniques allow for optimised representation, handling, and interpretation of classical linguistic data, thus offering quantum speedups, better accuracy, and improved security. 
Quantum optimisation techniques boost hyper parameter tuning process for NLP models that require optimisation over high dimensional parameter spaces. This integration of quantum algorithms with NLP models accelerates convergence and model efficiency. 
 Future technological advances are expected to overcome current hardware limitations, facilitating more robust solutions for complex NLP queries through QNLP.
\bibliographystyle{ACM-Reference-Format}
\bibliography{sources}
\end{document}